\documentclass[sigconf]{aamas} 

\usepackage{balance} 

\usepackage{amsmath}
\usepackage{booktabs,tabularx,multirow,makecell}
\usepackage{graphicx}
\usepackage{csquotes}
\usepackage{placeins}
\usepackage{longtable}
\usepackage{xurl}
\renewcommand\footnotetextcopyrightpermission[1]{} 
\setcopyright{none}
\acmConference[]{}{}{}
\acmBooktitle{}
\acmDOI{}
\acmISBN{}
\acmPrice{}
\acmYear{}
\copyrightyear{}

\title[AAMAS-2027 Formatting Instructions]{Beyond Preset Identities: Selective Stance Accommodation and Interaction Reorganisation in Generative Agent Societies}

\author{Hanzhong Zhang}
\affiliation{%
  \department{Department of Computer Science}
  \institution{University of Exeter}
  \city{Exeter}
  \country{United Kingdom}
}
\email{armihiabelliard@gmail.com}

\author{Siyang Song}
\affiliation{%
  \department{Department of Computer Science}
  \institution{University of Exeter}
  \city{Exeter}
  \country{United Kingdom}
}
\email{S.Song@exeter.ac.uk}

\author{Lu Liu}
\affiliation{%
  \department{Department of Computer Science}
  \institution{University of Exeter}
  \city{Exeter}
  \country{United Kingdom}
}
\email{L.Liu3@exeter.ac.uk}

\author{Jindong Wang}
\affiliation{%
  \department{Department of Data Science}
  \institution{William \& Mary}
  \city{Williamsburg}
  \state{Virginia}
  \country{United States}
}
\email{jdw@wm.edu}

\begin{abstract}
Generative agent societies simulate people with assigned roles, preferences and relationships. As agents exchange arguments and choose partners, they can revise their positions and reorganise discussion. Understanding these changes requires examining what they accept and how they continue to interact.

Stance-change scores and communication totals describe the extent of change. However, the same stance movement can preserve or reverse an assigned preference, and frequent communication can support either agreement or continuing disagreement. We therefore examine the content of changed positions and the exchanges that strengthen particular partnerships.

Using Computational Multi-Agent Society Experiments (CMASE), we combine stance measures, source evaluations and temporal networks with individual answers and messages. Study 1 compares seven conditions across ten GPT-4o runs per condition, with a separate interview collection covering four models. Study 2 follows one 75-step GPT-4o café simulation.

Environmental rational persuasion yields the largest mean stance departure ($1.30\pm0.08$ on a 7-point scale), whereas economic emotional persuasion yields the highest low-trust stance-shift rate ($17.3\%\pm11.2\%$, with standard deviations across runs). In the separate interviews, eight environmental agents shift from 7 to 6, acknowledge economic concerns and rate the source 3. Their partial acceptance preserves the assigned environmental preference. In the café, a pair with zero earlier exchanges becomes the most frequent final-phase partnership, with 25 messages. Its members develop coordination proposals while disputing their implementation. These results show that partial acceptance can coexist with low source trust, and sustained coordination with continuing disagreement.
\end{abstract}


\newcommand{\BibTeX}{\rm B\kern-.05em{\sc i\kern-.025em b}\kern-.08em\TeX}

\newcommand{\metric}[1]{\mathrm{#1}}
\newif\ifincludeappendix
\includeappendixtrue
\AtEndDocument{\nobalance}

\begin{document}




\maketitle 

\section{Introduction}\label{sec:introduction}
Generative agent societies use large language models (LLMs) to simulate participants who discuss shared concerns and choose conversation partners \citep{park2023generative,li2023camel,gao2024large}. Researchers initialise agents with roles and goals, supplemented by stance assignments or descriptions of other participants according to the study. We refer to this specification as the agent's \emph{preset identity}. Once interaction begins, agents encounter competing arguments and respond to particular speakers. A central question is how much they incorporate from those exchanges while maintaining an assigned preference or an ongoing discussion. This question concerns both the content of expressed positions and the relationships through which discussion continues.

Opinion-dynamics research examines consensus and disagreement in LLM-agent networks, while persuasion and persona studies investigate how language and context shape responses \citep{chuang2024opinion,breum2024persuasion,frisch2024interaction}. The distinction between agreement and continued engagement matters when interpreting their outcomes. For example, an absolute stance-change score combines movement towards and away from an argument, while a high cross-role message proportion leaves the particular conversational partners unspecified. We formulate two research questions to examine what these summaries mean in the underlying exchanges. \textbf{RQ1:} How do stance movement, its direction and reported trust differ across discursive conditions, and what do individual answers reveal about these differences? \textbf{RQ2:} How do conversational partnerships and interaction prominence change over time, and which exchanges accompany those changes?

To address these questions, we combine numerical comparisons with an analysis of the corresponding answers and messages. Both studies use Computational Multi-Agent Society Experiments (CMASE), which maintains memories and records actions, conversations and researcher interventions \citep{zhang2025cmase}. Study 1 models thirty residents discussing a proposed waste-incineration plant. Environmental and economic arguments provide opposing positions against which to measure movement and source evaluation. Study 2 examines ten agents in a café, where differentiated roles and a shared venue permit repeated encounters. The first study tests how agents revise an expressed position. The second examines how recurring partners develop and challenge proposals during continuing exchange.

Study 1 compares seven conditions with ten GPT-4o runs per condition, supplemented by a separate four-model interview collection. Environmental rational persuasion produces the largest mean departure, whereas economic emotional persuasion produces the highest low-trust shift rate. In the separate GPT-4o interviews, eight environmental agents approach the economic argument by one point, explicitly acknowledge its concerns and remain on the environmental side of the scale. We call this combination \emph{selective stance accommodation}: movement towards an opposing argument with both the assigned side and an acknowledged competing concern expressed in the answer. Low reported trust describes the source evaluation accompanying these cases. In Study 2, a previously inactive pair becomes the most frequent final-phase partnership while its members keep questioning coordination proposals. The two studies thus examine partial agreement in a position and continuing disagreement within a strengthening relationship.

The paper makes three contributions. \textbf{First}, it presents an evaluation procedure that separates movement magnitude, movement direction and source appraisal, and analyses partner changes through temporally ordered messages. \textbf{Second}, it shows that low source trust can accompany partial acceptance of an opposing argument, supported by repeated condition comparisons and concrete response patterns. \textbf{Third}, it demonstrates how a new partnership becomes prominent through reciprocal proposals and wider attention while disagreement over implementation continues. Together, these findings explain forms of adaptation that combine change with the persistence of a preference or discussion.

\section{Related Work}\label{sec:related}
\textbf{Generative agents and artificial societies.} LLM agents combine language generation with perception, memory and action \citep{xi2025rise}. Generative Agents and CAMEL establish persistent, role-conditioned interaction \citep{park2023generative,li2023camel}, whose use in modelling heterogeneous participants and collective behaviour is discussed in surveys \citep{gao2024large}. More specific settings include AgentGroupChat-V2's hierarchical task decomposition and adaptive collaboration \citep{gu2025agentgroupchat}, and CompeteAI's competition in a virtual market \citep{zhao2023competeai}. Across these settings, individual responses develop through repeated exchange. Our analysis follows this connection through expressed positions and the partners sustaining conversation.

\textbf{Opinion dynamics and persuasion.} Chuang et al. relate consensus and fragmentation to model tendencies and confirmation-bias prompts \citep{chuang2024opinion}. Within the same area, Ohagi examines polarisation under echo-chamber conditions \citep{ohagi2024polarization}. Breum et al. focus on persuasion, varying linguistic dimensions in synthetic dialogues and comparing agent and human judgements \citep{breum2024persuasion}. Building on these studies, our Study 1 examines the composition of a changed position. Specifically, magnitude, direction and reported trust describe how far an answer shifts, which side it approaches and how the agent evaluates the source. Exposure histories further relate that report to preceding communication.

\textbf{Persona configuration and interactional consistency.} Persona prompting and role-play specify behavioural expectations \citep{white2023prompt,njifenjou2024role}. RoleLLM and CharacterEval assess the resulting responses \citep{wang2024rolellm,tu2024charactereval}, while Tseng et al. distinguish model personas from user personalisation \citep{tseng2024two}. Other work examines configuration during generation: in-context impersonation studies assigned personas and model tendencies \citep{salewski2023context}, and inner-monologue methods develop topic guidance and proactive questioning \citep{zhou2024think}. At the population level, Frisch and Giulianelli study personality consistency and linguistic alignment \citep{frisch2024interaction}. Persona Alchemy grounds stakeholder agents in Social Cognitive Theory and varies the reliability of contradictory information \citep{kim2025persona}. This relation between configuration and subsequent information is closest to our question of what agents incorporate within a retained preference or continuing partnership.

\textbf{Evaluation and social simulation.} AgentBench, AgentBoard, SOTOPIA and Avalon-based evaluation assess tasks, trajectories and social interaction \citep{liu2023agentbench,chang2024agentboard,zhou2023sotopia,light2023text}. Argyle et al. also examine demographic conditioning for synthetic response samples \citep{argyle2023out}. Interpreting such outputs calls for explicit assumptions. Social-simulation methodology connects those assumptions and explanatory purpose to local interaction and aggregate patterns \citep{gilbert2005simulation,epstein2008why}. The Overview, Design concepts and Details (ODD) protocol structures model purpose, scheduling, initialisation and evaluation \citep{grimm2020odd}. Following this emphasis, our procedure retains each result's reference and statistical unit before linking it to source answers or messages.

\textbf{Interaction processes and situated interpretation.} Research on socially enactive cognition relates an agent's processing to its ongoing social interaction \citep{kahl2023intertwining}. At a broader level, Tsvetkova et al. examine configurations of humans and machines as social systems \citep{tsvetkova2024new}. To study these relations in our café case, we follow addressees, recurring formulations and later elaborations chronologically. Network measures locate changes in attention and partnerships, while dialogue identifies the content around which those partnerships develop. On this basis, the ethnographic account in Section~\ref{app:ethnography_expanded} interprets temporary groupings and discursive stabilisation.

\section{Methodology}\label{sec:design}
\subsection{Study design}\label{sec:references}
We implement both studies in CMASE, which constructs a shared environment from a map and natural-language descriptions \citep{zhang2025cmase}. At each update round, called a simulation step, an agent receives its configured identity, current observations and retrieved memories before selecting an action. The resulting conversation record stores the speaker, addressee and message. Event entries separately record researcher-specified situations, so the log distinguishes supplied events from addressed responses. Researcher participation uses this same recorded environment.

\begin{table*}[t]
\caption{Common analytical structure and the evidence used in each study.}
\label{tab:study_map}
\small
\begin{tabularx}{\textwidth}{>{\raggedright\arraybackslash}p{.16\textwidth}>{\raggedright\arraybackslash}p{.23\textwidth}>{\raggedright\arraybackslash}p{.24\textwidth}>{\raggedright\arraybackslash}X}
\toprule
Analysis & Material & Reference and outcomes & Contribution \\\midrule
Study 1: repeated conditions & Seven conditions, ten 30-agent GPT-4o runs per condition & Preset stance, neutral-subgroup direction, departure and low-trust shift rate & Distinguishes intervention outcome profiles \\
Study 1: individual follow-up & Four GPT-4o endpoint files, 30 agents each, sixteen files spanning models and conditions in the supplement & Explicit stance/trust responses, signed movement and source appraisal & Explains the composition and expressed meaning of the profiles \\
Study 2: relational process & One 75-step café record, ten agents and an embedded researcher & Initial partner descriptions, early strength ranks, addressed chats and recurring pairs & Traces reorganisation of participation and coordination \\\bottomrule
\end{tabularx}
\end{table*}

Table~\ref{tab:study_map} summarises the evidence for each question. Repeated GPT-4o runs estimate differences between the residential conditions. Separate interviews compare individual positions and source evaluations across four models. In the café, the complete message sequence measures changing partnerships and records what the partners discuss. These complementary analyses test whether an agent's expressed movement amounts to accepting an opposing concern and whether a frequent relationship sustains the negotiation of a proposal.

\subsection{Analysis procedure}\label{sec:simulation}
For Study 1, we compare each final stance with its preset score and calculate movement towards or away from the intervention. The final score determines whether the answer remains on its assigned side of the neutral midpoint. We then read the answer for an explicit acknowledgement of the competing concern. Reported trust and its explanation separately measure \emph{source appraisal}, the agent's stated evaluation of the new resident. An \emph{endpoint interview} elicits these reports after a simulation. Run summaries support condition comparisons, while complete interview answers support the response classifications.

For Study 2, we parse each message's step, sender and recipient, then construct directed networks with message counts as edge weights. An agent's total sent and received messages measure \emph{conversational prominence}. Pair counts measure repeated contact, and five-step recipient shares measure shorter concentrations of attention. We then read the associated turns for coordination proposals, replies and objections. This distinguishes a frequently contacted participant from a partnership that repeatedly develops a proposal. Event descriptions and other actions explain the situations in which those exchanges occur.

\section{Study 1: Stance Adaptation}\label{sec:study1}
\subsection{Experimental setup}\label{sec:study1_protocol}
Using CMASE, we simulate a residential community discussing whether to build a waste-incineration plant. Because the issue combines environmental and economic concerns, the setting supports comparisons of expressed stance and evaluations of an argument's source. Thirty simulated residents interact for 21 steps, with ten assigned to each of three groups: Environmental Advocates, Economic Growth Supporters and Neutral Residents. Within this community, researcher-written statements are delivered as the speech of a new resident.

Four interventions combine environmental (Env) or economic (Eco) orientation with rational persuasion (RP) or emotional persuasion (EP): Env-RP, Env-EP, Eco-RP and Eco-EP. RP presents facts and logical arguments, whereas EP invokes fear, empathy, moral concern or livelihood pressure. In addition, three comparison conditions vary the available interaction. Non-Social omits peer communication and researcher intervention. Natural Social retains community exchange without targeted persuasion, while Isolated presents the Eco-EP statements outside the interactive community. Section~\ref{app:population} gives the population and delivery details.

Stance is reported on a 1--7 scale from economic growth to environmental protection, with 4 neutral. Reported trust in the new resident presenting the intervention uses a separate 1--7 scale. For agent $i$, let $p_i\in\{1,4,7\}$ be its preset stance, $s_i$ its final stance and $t_i$ its reported trust. Let $\mathcal N_0$ denote the initially neutral agents and $N=30$. For a movement threshold $\delta$ and trust threshold $\tau$, we calculate three complementary measures:
\begin{align}
\metric{DVB}&=\frac{1}{|\mathcal N_0|}\sum_{i\in\mathcal N_0}s_i-4,\label{eq:dvb}\\
\metric{PS}&=\frac1N\sum_i |s_i-p_i|,\label{eq:ps}\\
\metric{LTSS}_{\delta,\tau}&=\frac{100}{N}\sum_i\mathbf1\{|s_i-p_i|\geq\delta,\ t_i\leq\tau\}.\label{eq:ltss}
\end{align}
Discursive Value Bias (DVB) measures final direction in the initially neutral group, while Persuasion Sensitivity (PS) measures mean absolute departure from the preset stance. The Low-Trust Stance-Shift (LTSS) rate additionally records the percentage of agents whose movement meets the threshold and whose reported trust is low. We use the reported one-point rule, $(\delta,\tau)=(1,3)$, for the repeated results. A joint case therefore meets both criteria. For the separate interview analysis, we also retain each case's movement direction and final position.

The repeated comparison contains ten GPT-4o runs per condition. We calculate means and sample standard deviations, then use the run as the unit for two-sided Welch tests. Individual analysis uses separate final interviews, with thirty agents per model and intervention. A fixed rule extracts explicit ratings, preserves fractions and removes scale denominators and list numbers. Ambiguous ratings remain unscored, with details in Sections~\ref{app:contrasts} and~\ref{app:crossmodel}. Thus, run summaries estimate condition differences, while interviews identify individual combinations of stance and source appraisal.

\paragraph{Response classification.} For orientation $c$ ($c=-1$ for economic and $c=+1$ for environmental), $d_i=c(s_i-p_i)$ gives signed movement towards the intervention. We classify an answer as selective stance accommodation when an initially opposing agent moves towards the intervention by at least one point, remains on its assigned side of 4 and explicitly acknowledges a concern advanced by the intervention. The numerical conditions are $c(p_i-4)<0$, $d_i\geq1$ and $(p_i-4)(s_i-4)>0$. For example, 7 to 6 under economic persuasion meets these conditions when the answer incorporates an economic concern. An unchanged score, a move away from the argument or a reversal to its side belongs to a different response category. Trust is coded separately, so accommodation can occur at either high or low source ratings.

\subsection{Intervention materials}\label{sec:materials}
The interventions differ in their language as well as their orientation. The collection contains 80 statements, each with three LLM ratings on Emotional Pressure, Threat Framing, Content Details and Moral Valence. These descriptors assess urgency, threatened consequences, informational detail and moral framing, respectively. EP receives higher ratings for pressure, threat and moral framing, whereas RP receives higher ratings for informational detail (Table~\ref{tab:semantic_main}). The environmental and economic materials differ less on these dimensions (Section~\ref{app:materials}). Thus, the material analysis confirms the rhetorical contrast being compared in the response analysis.

\begin{table}[t]
\caption{Mean language ratings for emotional (EP) and rational (RP) intervention statements. The four descriptors characterise the supplied materials.}\label{tab:semantic_main}
\small\begin{tabular}{lrr}\toprule
Descriptor & EP & RP\\\midrule
Emotional Pressure & 4.31 & 1.29\\
Threat Framing & 3.39 & 1.61\\
Content Details & 1.36 & 2.74\\
Moral Valence & 4.34 & 1.67\\\bottomrule
\end{tabular}\end{table}

This rhetorical contrast helps explain the questions asked of each response. An agent may acknowledge employment needs but criticise a divisive appeal. We therefore examine both the position expressed and the reasons given for the source rating, keeping the argument's orientation and presentation together as the experimental condition.

\subsection{Repeated-run results}\label{sec:study1_results}
\begin{figure}[t]
\centering\includegraphics[width=\linewidth]{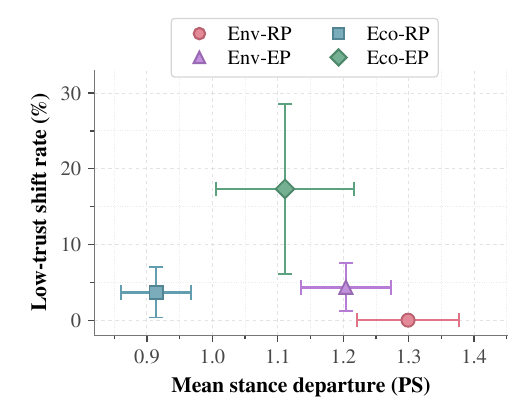}
\caption{Stance outcomes for the four discursive conditions. Horizontal and vertical bars show sample standard deviations across runs. The two axes compare mean absolute departure with the frequency of low-trust shifts.}
\label{fig:profiles}
\Description{Four intervention conditions positioned by mean stance departure and low-trust shift rate, with standard deviation bars.}
\end{figure}

Env-RP has the largest mean PS ($1.30$) across the seven conditions in Table~\ref{tab:repeat}, whereas Eco-EP has the highest reported LTSS ($17.3\%$). In particular, Eco-EP exceeds Eco-RP by 13.66 percentage points (95\% confidence interval $[5.48,21.84]$, $p=.0038$) and Non-Social by 16.35 points (95\% CI $[8.30,24.40]$, $p=.0012$). As Figure~\ref{fig:profiles} shows, the greatest average departure and the most frequent low-trust shifts occur under different interventions.

\begin{table*}[t]
\caption{Repeated Study 1 outcomes: ten runs per condition. Values are means $\pm$ sample standard deviations calculated from the recorded per-run summaries. DVB refers to initially neutral agents. The other summaries refer to the community. The scoring procedure and run comparisons are described in Section~\ref{app:contrasts}.}
\label{tab:repeat}\label{tab:main_replication}
\small
\begin{tabular}{lrrrr}\toprule
Condition & DVB & PS & LTSS (\%) & Reported trust \\\midrule
Non-Social & $0.69\pm0.07$ & $0.84\pm0.04$ & $1.0\pm1.6$ & $5.05\pm0.10$ \\
Natural Social & $0.66\pm0.08$ & $0.97\pm0.06$ & $4.0\pm3.1$ & $4.46\pm0.12$ \\
Isolated & $0.42\pm0.10$ & $0.66\pm0.09$ & $7.0\pm2.5$ & $5.04\pm0.09$ \\
Eco-EP & $0.19\pm0.10$ & $1.11\pm0.11$ & $17.3\pm11.2$ & $4.82\pm0.52$ \\
Eco-RP & $0.28\pm0.09$ & $0.91\pm0.05$ & $3.7\pm3.3$ & $5.61\pm0.35$ \\
Env-EP & $1.32\pm0.15$ & $1.20\pm0.07$ & $4.3\pm3.2$ & $5.48\pm0.33$ \\
Env-RP & $1.70\pm0.09$ & $1.30\pm0.08$ & $0.0\pm0.0$ & $6.10\pm0.27$ \\
\bottomrule\end{tabular}
\end{table*}

The initially neutral group finishes on the environmental side in every condition, including Non-Social and Natural Social. Positive PS in those two conditions also shows movement without targeted persuasion. Under the same Eco-EP material, the interactive community has greater mean departure than Isolated ($1.11$ vs $0.66$) and a higher LTSS rate ($17.3\%$ vs $7.0\%$). The LTSS difference is 10.33 percentage points (95\% CI $[2.23,18.43]$, $p=.0176$). This comparison evaluates the combined community protocol, in which exchange and accumulated context accompany the intervention. The separate interviews show how agents express partial acceptance and source reservations within individual answers.

\subsection{Individual response analysis}\label{sec:study1_cases}
The separate GPT-4o interviews show that most Eco-EP low-trust shifts are small adjustments towards an opposing argument. Across the four condition files, we examine all 120 stance/trust pairs. Table~\ref{tab:followup} reports movement and low trust separately, their intersection, and its direction. Eco-EP contains 20 changed positions, 19 low source ratings and 11 joint cases. Eight joint cases approach the economic intervention, while three move in the opposite direction. Ten of the eleven shift by exactly one point. Direction and the final preference therefore distinguish partial acceptance from other changes contributing to the same joint rate.

\begin{table}[t]
\caption{Individual interview results: one 30-agent file per GPT-4o condition. Changed means $|s_i-p_i|\geq1$. Low means $t_i\leq3$. Toward and Away partition the joint cases by intervention direction.}
\label{tab:followup}
\small\setlength{\tabcolsep}{3pt}
\begin{tabular}{lrrrrr}\toprule
Condition & Changed & Low & Joint & Toward & Away \\\midrule
Env-RP & 20 & 0 & 0 & 0 & 0 \\
Env-EP & 21 & 2 & 2 & 2 & 0 \\
Eco-RP & 14 & 5 & 1 & 1 & 0 \\
Eco-EP & 20 & 19 & 11 & 8 & 3 \\
\bottomrule\end{tabular}
\end{table}

All eight target-directed joint cases satisfy the criteria for selective stance accommodation. These simulated Environmental Advocates move from 7 to 6, acknowledge an economic concern and give the source a trust rating of 3. Two examples are Chelsea Warnes and Vickie Houck. Chelsea prioritises sustainable jobs and environmental protection but criticises the source's confrontational presentation and lack of a balanced proposal. Vickie likewise acknowledges the urgency of jobs while criticising a divisive approach. The responses accept part of the economic argument without reversing the assigned environmental preference.

Chelsea's paired answers show how content acceptance and source criticism coexist. In her stance interview, the Environmental Advocate states, ``While acknowledging the need for economic provisions, my priority remains towards environmental protection and identifying sustainable jobs that do not compromise community health.'' Her source evaluation calls the concerns ``valid and important'' and cites the ``confrontational tone and lack of a proposed balanced solution''. Thus, the low rating accompanies explicit acceptance of the economic concern, while the final stance continues to prioritise environmental protection. Section~\ref{app:cases} gives all four examples in full.

The comparison cases clarify the classification. Ronnie Brooks, a simulated Neutral Resident in the same file, acknowledges economic needs and criticises one-sided presentation, yet remains at 4. His acknowledgement produces no measured stance accommodation. Economic Growth Supporter Alice Sanchez moves from 1 to 3 and rates the source 4, which is movement away from the economic intervention. Across all thirty Eco-EP source explanations, a dictionary matches presentation terms such as \texttt{confrontational}, \texttt{divisive} or \texttt{one-sided} in eight answers, compared with one under Env-EP and zero in the rational conditions. Sections~\ref{app:cases} and~\ref{app:appraisal} give the answers and coding rules.

These combinations are also distributed differently across the preset groups. Among Environmental Advocates, nine of ten move and nine report low trust, with eight meeting both criteria. All ten Economic Growth Supporters move, but just two report low trust. Conversely, eight Neutral Residents report low trust and one moves. Thus, movement is widespread in the economic group and low trust in the neutral group, while their overlap is concentrated in small environmental adjustments towards the economic position.

\subsection{Exposure analysis}\label{sec:exposure}
The relationship between cross-group exposure and stance movement differs across the two economic conditions. Exposure is the share of incoming messages initiated outside the recipient's preset group. In the Eco-RP records, greater exposure accompanies larger shifts ($\rho=.488$, nominal $p=.0063$, 30 agents). Changed agents have mean exposure .774, compared with .519 for unchanged agents. In Eco-EP, however, the means are .640 and .657, respectively, and the rank association is near zero ($\rho=-.021$). All three agents with zero cross-group exposure change, as do three of six receiving exclusively cross-group messages. Thus, the Eco-EP changes span the exposure range. This contrast gives no uniform relationship between exposure and movement across the economic conditions. Section~\ref{app:exposure} reports all four trajectories.

\subsection{Cross-model comparison}\label{sec:model_profiles}
The four-model comparison shows that larger movement need not accompany lower source ratings. It covers GPT-4o, Gemini-2.5-Flash, Llama-3.1-8B and Qwen-3-8B under all four interventions, with one endpoint file per model and condition. PS ranges from .13--.20 for Gemini, .90--1.33 for GPT-4o, 1.95--2.00 for Llama and 1.29--1.45 for Qwen under the same extraction rule. Table~\ref{tab:model_profiles_main} compares these ranges with reported trust and LTSS. Section~\ref{app:crossmodel} gives all sixteen condition rows and valid denominators.

\begin{table}[t]
\caption{Ranges across the four endpoint condition means for each model. LTSS uses a one-point movement threshold and valid paired answers. Each range runs from the minimum to the maximum observed condition mean.}\label{tab:model_profiles_main}
\small\setlength{\tabcolsep}{3pt}
\begin{tabular}{lrrr}\toprule
Model & PS & Trust & LTSS (\%)\\\midrule
GPT-4o & 0.90--1.33 & 3.37--5.30 & 0.0--36.7 \\
Gemini-2.5-Flash & 0.13--0.20 & 3.67--3.87 & 3.3--13.3 \\
Llama-3.1-8B & 1.95--2.00 & 4.77--5.21 & 0.0--7.1 \\
Qwen-3-8B & 1.29--1.45 & 4.87--5.20 & 0.0--3.6 \\
\bottomrule\end{tabular}\end{table}

The models exhibit different combinations of movement and source ratings. Gemini has small departures with joint low-trust cases in every condition. By contrast, Llama has larger departures and higher source ratings: three conditions have zero joint cases, while Eco-RP has two among 28 scored pairs. Qwen likewise has few joint cases, with one under Env-RP and zero elsewhere. GPT-4o's largest joint group occurs under Eco-EP. The full table retains all four conditions per model.

The summaries use valid-answer counts as denominators, with ambiguous answers recorded as unscored. Across the files, Gemini's small departures coexist with lower source ratings, while Llama's larger departures coexist with higher ratings. The repeated GPT-4o experiment separately estimates differences between conditions.

\subsection{Group-level results}\label{sec:legacy_groups}
\begin{figure}[t]
\centering\includegraphics[width=\linewidth]{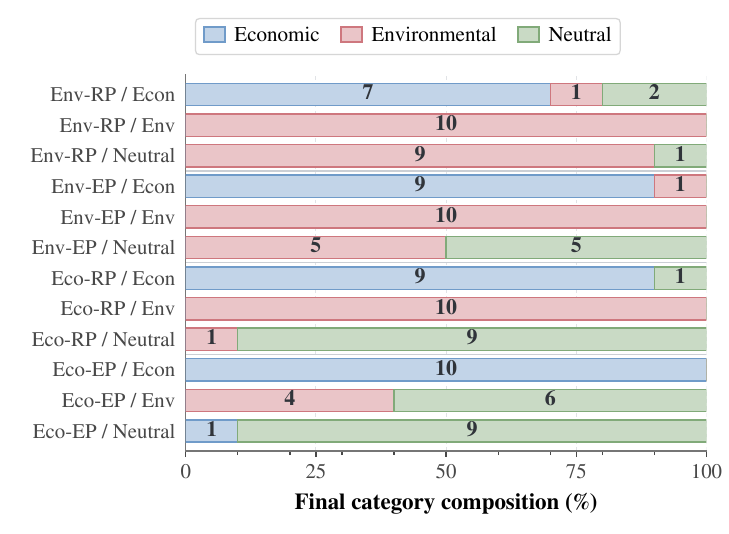}
\caption{Categorical stance transitions: final group composition within each preset group (Econ, Env, Neutral), ten agents per row. The four conditions remain separate. The categorical response table supplies the counts. Supplementary plots show the corresponding source ratings and communication.}\label{fig:group_transitions}
\Description{Twelve stacked bars show final categorical stances for all preset-group and intervention combinations.}
\end{figure}

A separate categorical protocol measures changes between the three stance groups and trust on a 10-point scale. Its 120-agent table covers all four conditions (Figure~\ref{fig:group_transitions}). Env-RP changes 12 of 30 category assignments, Env-EP changes 6, Eco-RP changes 2 and Eco-EP changes 7. Their mean trust scores are 7.87, 6.83, 6.57 and 6.10, respectively. Here participation counts residents present in the central square, while the affect score averages the summed valence, arousal and dominance category values of its members. Env-EP sustains greater square occupancy, yet Env-RP produces more category changes. Section~\ref{app:original_figures} defines these measures and shows their trajectories alongside the categorical results.

\section{Study 2: Interaction Dynamics}\label{sec:study2}

\begin{figure*}[t]
\centering\includegraphics[width=0.98\textwidth]{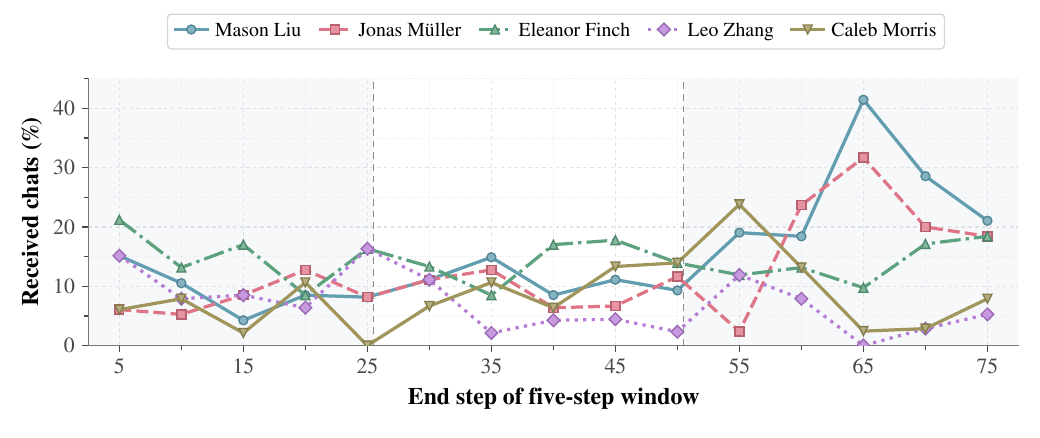}
\caption{Share of café agent chats received by five named agents in successive five-step windows. Dashed lines separate early interaction, extended researcher participation and event-related interaction. Mason is staff, Jonas a tourist, Eleanor the owner, Leo a student and Caleb a regular customer. Each share uses all retained agent chats in its window. Data cover all ten agents.}
\label{fig:attention}
\Description{Recipient-share trajectories for Mason, Jonas, Eleanor, Leo, and Caleb over the full seventy-five-step record.}
\end{figure*}
\subsection{Experimental setup}\label{sec:study2_methods}
To examine the relational aspect of adaptation, we use CMASE to simulate repeated encounters among ten agents in a café. The shared venue brings together an owner, staff, regular customers, students, tourists and a cleaner, each with goals and descriptions of other participants. Because these roles interact in the same setting, we can follow which partners become recurrent as conversations continue. The analysis uses one complete 75-step GPT-4o simulation, including a researcher participant who enters discussion through the framework's control interface.

Specifically, the cast comprises owner Eleanor Finch, staff Mason Liu and Jade Herrera, regulars Caleb Morris and Nina Takahashi, students Ava Ramires and Leo Zhang, tourists Emily Novak and Jonas Müller, and cleaner Marisol Reyes. These names identify simulated agents throughout the figures and excerpts. Each agent initially describes selected others, with directed descriptions that can differ between members of a pair.

We divide this record into three consecutive phases: initial interaction (P1, Steps~1--25), extended researcher participation (P2, Steps~26--50) and event-related interaction (P3, Steps~51--75). Although chats involving the researcher occur throughout, the phases follow the protocol's changes in participation and event inputs. Confrontation scenes and response instructions enter at Steps~51--52, followed by a silence-related event at Step~56. Section~\ref{app:population} supplies the full role configuration and schedule.

For each phase, we construct a directed weighted graph from the addressed chats. Each non-empty \texttt{sender: chat with receiver: message} entry contributes once to the weight $w_{ij,p}$ from sender $i$ to recipient $j$ in phase $p$. The primary graph contains the ten simulated agents, supplemented by a second graph that includes the researcher. Let $r_i$ denote an agent's role and $\mathcal P_i$ the other agents named in its initial descriptions. Using these graphs and descriptions, we calculate interaction strength, Cross-Group Interaction across roles ($\metric{CGI}^{\mathrm{role}}$) and the share of messages to unlisted partners:
\begin{align}
q_{i,p}&=\sum_j(w_{ij,p}+w_{ji,p}),\label{eq:strength}\\
\metric{CGI}^{\mathrm{role}}_p&=\frac{\sum_{ij}w_{ij,p}\mathbf1[r_i\ne r_j]}{\sum_{ij}w_{ij,p}},\label{eq:cgi}\\
U_p&=\frac{\sum_{ij}w_{ij,p}\mathbf1[j\notin\mathcal P_i]}{\sum_{ij}w_{ij,p}}.\label{eq:unlisted}
\end{align}
Interaction strength $q$ counts all chats an agent sends and receives, measuring conversational prominence. Role mixing instead records communication across categories, while $U$ records messages to people absent from the sender's initial descriptions. To track relative prominence over time, rank stability, $\rho_{1p}$, compares phase strengths with Phase~1 using Spearman correlation and average ranks for ties. At a finer temporal scale, recipient share is the fraction of messages addressed to an agent in a five-step window. Pair counts sum both directions of exchange. We display the two most frequent pairs from each phase and retain their union, with all possible pairs in the data (Section~\ref{app:network}).

We assess coordination in the messages exchanged by recurring pairs. A coordination proposal suggests joint activity, shared goals or subsequent steps. Negotiated coordination requires reciprocal exchange across steps, a proposal that is continued or developed in later turns, and an explicit qualification or objection concerning its implementation. A high message count alone measures contact frequency. We apply these reading criteria to the complete sequence surrounding a partnership, using fixed word lists to count coordination vocabulary and explicit proposals throughout the record. Sections~\ref{app:windows} and~\ref{app:sequences} give the post-hoc dictionaries and the corresponding turns.

\subsection{Network results}\label{sec:study2_results}
Communication shifts towards new partners while cross-role exchange remains common. Of the 635 non-empty café chats, 183 of 214 in P1 and 183 of 194 in P3 cross occupational roles. Over these phases, the share addressed to partners absent from initial descriptions rises from 25.2\% to 43.8\%, while interaction-strength ranks correlate at $-.396$ with P1 (Table~\ref{tab:network}). The changing ranks and partners reveal a redistribution of prominence within a network that already communicates extensively across roles.

\begin{table}[t]
\caption{Phase-level interaction summaries. $I$ counts agent chats. $U$ is the share directed to partners absent from the sender's initial descriptions. $\rho_{1p}$ compares strength ranks with Phase~1.}
\label{tab:network}
\small\begin{tabular}{lrrrr}\toprule
Phase & $I$ & CGI$^{\mathrm{role}}$ & $U$ & $\rho_{1p}$\\\midrule
1 (1--25) & 214 & .855 & .252 & 1.000\\
2 (26--50) & 227 & .846 & .238 & .245\\
3 (51--75) & 194 & .943 & .438 & $-.396$\\\bottomrule
\end{tabular}
\end{table}

The clearest change involves staff member Mason Liu and tourist Jonas Müller. Mason moves from strength 42 and rank 6 to strength 69 and rank 1, while Jonas moves from strength 40 and rank 7 to strength 60 and rank 2. In comparison, owner Eleanor Finch moves from rank 1 to rank 4, and student Leo Zhang from rank 3 to tied rank 8.5. Mason and Jonas have zero direct chats in the first two phases and are absent from each other's initial descriptions. Nevertheless, their 25 Phase~3 chats form the most frequent pair, constituting 12.9\% of all agent chats.

At the same time, active partnerships expand from 24 in P1 and 20 in P2 to 33 in P3, among 45 possible pairs. Messages to unlisted partners increase from 54 and 54 to 85. The new frequent pair therefore develops alongside a broader set of active ties. Its prominence also survives inclusion of researcher messages: Mason remains the strongest final-phase node at 80 (Section~\ref{app:network}). Table~\ref{tab:pair_paths} compares the new pair with earlier relationships.

\begin{table}[t]
\caption{The two most frequent pairs in each phase, shown across all three phases. The union contains six pairs. The appendix archive retains all 45 possible pairs. Names are abbreviated to first names.}\label{tab:pair_paths}
\small\begin{tabular}{lrrr}\toprule
Pair & P1 & P2 & P3\\\midrule
Marisol and Nina & 19 & 10 & 6\\
Eleanor and Mason & 18 & 14 & 8\\
Ava and Emily & 10 & 16 & 1\\
Caleb and Mason & 3 & 15 & 10\\
Jonas and Mason & 0 & 0 & 25\\
Ava and Jonas & 9 & 12 & 17\\\bottomrule
\end{tabular}\end{table}

Several earlier ties persist as their frequencies change. Exchanges decrease for Marisol and Nina and for Eleanor and Mason. Caleb and Mason peak in the middle phase but remain active later, whereas Ava and Jonas increase throughout. Ava and Emily also peak in the middle phase, then exchange one final-phase message. In contrast, Jonas and Mason become the most frequent pair in the final phase after having no direct exchanges earlier.

\subsection{Early interaction}\label{sec:early_groups}
Before the final-phase partnership forms, agents repeatedly discuss authenticity, influence, tradition and café management. Student Leo asks tourist Jonas about indirect expression and fellow student Ava about silence and openness. These questions recur through particular partners even as each participant joins other conversations. Consequently, early interaction already combines repeated relationships with movement between topics and speakers.

Researcher participation broadens the discussion to order and identity. At Step~26, the researcher asks Eleanor whether order comes from particular speakers or spatial arrangement. Eleanor responds at Step~29 by asking whether an observer reshapes the space. Later exchanges develop labels, multiple selves, contradiction and coexistence. Participants thus turn the researcher's questions into topics they can question and extend. Maffesoli's provisional groupings describe these temporary associations maintained through recurrent topics and mutual attention \citep{maffesoli2022}.

\subsection{Coordination analysis}\label{sec:study2_process}
The confrontation draws attention to Leo and Caleb, but that attention declines as coordination talk becomes frequent. At Step~52, Emily asks Leo about silencing voices, Ava questions his challenge, and Leo asks Caleb about conflict and coexistence. Leo and Caleb receive 15 of 42 chats during Steps~51--55 (35.7\%), falling to 8 of 114 during Steps~61--75 (7.0\%). Mason and Jonas receive 62 in the latter window (54.4\%). Figure~\ref{fig:attention} shows both episodes across the full series. Attention to the supplied conflict and attention to later coordination differ in their recipients and duration.

Mason and Jonas become prominent through reciprocal discussion and replies from the wider café. Mason addresses Jonas in 14 of the last 15 steps, and Jonas addresses Mason in ten. Their 24 mutual messages accompany 38 received from seven other agents. Their proposals also develop across turns: Mason connects adaptability with shared goals at Step~67, adds structured collaboration at Step~69, and Jonas proposes actionable steps. Subsequent exchanges discuss frameworks, metrics and documentation. Repeated contact therefore sustains the development of a shared coordination problem.

However, the recurring exchange also contains objections. Jonas questions whether structure could inhibit nuanced progress and contrasts its clarity with the opportunities afforded by adaptation. Coordination vocabulary appears in 1/214, 2/227 and 37/194 messages across the phases, with fourteen final-phase messages also containing explicit proposals. The increase marks a shift towards coordination talk, while the dialogue shows what remains disputed. Mason and Jonas continue proposing shared arrangements as Jonas challenges their implementation. This combination satisfies the criteria for negotiated coordination, with recurrence and objection both visible in the sequence.

For example, at Step~67 Mason tells Jonas, ``Jonas, fostering resilience requires balancing adaptability with shared goals. Let's suggest frameworks to align these values as a group.'' Later turns propose actionable steps and documentation while Jonas questions rigidity. Thus, agreement on the need to coordinate sustains exchange even while the means of coordination remain contested. The new partnership strengthens through continuing discussion of this unresolved problem.

\section{Discussion}\label{sec:discussion}
\textbf{Partial acceptance can coexist with source distrust.} The eight environmental agents approach the opposing argument while continuing to prioritise environmental protection and rating the source 3. Their answers explain this combination by accepting economic needs and questioning presentation or balance. Selective stance accommodation therefore distinguishes three decisions within a response: how much to change, which side to support and how to evaluate the speaker. The examples of unchanged neutrality and movement away from the intervention show that acknowledgement and low trust alone do not meet its definition.

\textbf{A partnership can strengthen while disagreement continues.} Mason and Jonas become the most frequent pair while debating how coordination should work. Earlier relationships persist at different frequencies, so the change redistributes attention within the café's existing roles. Repeated coordination expressions organise subsequent proposals and objections, which we call \emph{discursive stabilisation}. Deterritorialisation describes the redirection of earlier alignments, while reterritorialisation describes the new partners and formulations around which exchange gathers \citep{deleuze2009,deleuze2013}. These concepts interpret the sequence through which a new partnership becomes a recurrent focus of discussion.

\textbf{Adaptation involves more than the amount of change.} Env-RP produces the greatest mean departure, whereas Eco-EP produces the most frequent low-trust shifts. Separating direction from magnitude then shows that eight of the eleven joint interview cases approach the economic argument and three move away. Similarly, the café's extensive cross-role communication coexists with a new most frequent pair. The two studies address different consequences of continued interaction: which concerns enter a position and which proposals keep partners engaged. Both show why aggregate activity and agreement must be examined separately.

\textbf{The findings guide evaluation.} A persuasion evaluation should distinguish partial accommodation, preference reversal and unchanged responses using direction, final position and answer content, then compare source ratings across those categories. A relational evaluation should compare partner recurrence with the proposals and objections expressed across successive messages. These distinctions yield directly inspectable outcomes, such as a one-point shift with a continuing environmental preference or a high-frequency partnership that keeps debating implementation. The accompanying scoring and message-processing scripts implement these comparisons.

\textbf{Response content matters alongside contact composition.} Eco-EP raises LTSS by 10.33 percentage points relative to isolated delivery of the same statements. Within its exposure records, however, changed and unchanged agents receive comparable cross-group shares. The community-level difference and the within-trajectory association therefore answer different questions. Together with the interview responses, these findings suggest that the content an agent accepts and its appraisal of the speaker matter alongside the amount of contact across groups. In the café, similarly, repeated contact becomes a coordination partnership through the proposals and objections expressed within it.

\section{Conclusion}\label{sec:conclusion}
We presented an evaluation method for expressed positions and conversational partnerships in generative agent societies. Across repeated conditions, the greatest mean stance departure and the highest low-trust shift rate occur under different interventions. Eight environmental agents in the separate interviews partially accept an economic argument while maintaining an environmental preference and low source ratings. In the café, a previously inactive pair becomes the most frequent partnership through reciprocal coordination proposals, replies from others and continuing objections. These findings show that adaptation can preserve a preference or sustain a relationship without requiring full agreement.

\bibliographystyle{ACM-Reference-Format} 
\bibliography{sn-bibliography}

@book{deleuze2009,
  author = {Deleuze, Gilles and Guattari, F{\'e}lix},
  title			= "Anti-Oedipus: Capitalism and Schizophrenia",
  address		= "London",
  publisher		= "Penguin Classics",
  year			= "2009",
  isbn={9780143105824}
}

@book{deleuze2013,
  author = {Deleuze, Gilles and Guattari, F{\'e}lix},
  title			= "A Thousand Plateaus",
  address		= "London",
  publisher		= "Bloomsbury Academic",
  year			= "2013",
  isbn={9781780935379},
  url={https://www.bloomsbury.com/uk/thousand-plateaus-9781780935379/}
}

@book{maffesoli2022,
  author = {Maffesoli, Michel},
  title			= "The time of the tribes: The decline of individualism in mass society",
  address		= "Shanghai",
  publisher		= "Shanghai People's Publishing House",
  year			= "2022",
  isbn={9787208175396},
  note={Chinese edition. Translated by Yibing Xu}
}

@inproceedings{park2023generative,
  title={Generative agents: Interactive simulacra of human behavior},
  author={Park, Joon Sung and O'Brien, Joseph and Cai, Carrie Jun and Morris, Meredith Ringel and Liang, Percy and Bernstein, Michael S},
  booktitle = {Proceedings of the 36th Annual ACM Symposium on User Interface Software and Technology},
    doi = {10.1145/3586183.3606763},
    publisher = {Association for Computing Machinery},
  pages={1--22},
  year={2023},
  address={New York, NY, USA}
}

@article{tsvetkova2024new,
  title={A new sociology of humans and machines},
  author={Tsvetkova, Milena and Yasseri, Taha and Pescetelli, Niccolo and Werner, Tobias},
  journal={Nature Human Behaviour},
  volume={8},
  number={10},
  pages={1864--1876},
  year={2024},
  publisher={Nature Publishing Group UK London},
  doi={10.1038/s41562-024-02001-8},
  url={https://www.nature.com/articles/s41562-024-02001-8}
}

@inproceedings{zhao2023competeai,
 title={{CompeteAI}: Understanding the Competition Dynamics of Large Language Model-based Agents},
 author={Zhao, Qinlin and Wang, Jindong and Zhang, Yixuan and Jin, Yiqiao and Zhu, Kaijie and Chen, Hao and Xie, Xing},
 booktitle={Proceedings of the 41st International Conference on Machine Learning},
 series={Proceedings of Machine Learning Research}, volume={235},
 pages={61092--61107}, year={2024}, publisher={PMLR}, address={Vienna, Austria},
 url={https://proceedings.mlr.press/v235/zhao24q.html}
}

@article{xi2025rise,
  title={The rise and potential of large language model based agents: A survey},
  author={Xi, Zhiheng and Chen, Wenxiang and Guo, Xin and He, Wei and Ding, Yiwen and Hong, Boyang and Zhang, Ming and Wang, Junzhe and Jin, Senjie and Zhou, Enyu and others},
  journal={Science China Information Sciences},
  volume={68},
  number={2},
  pages={121101},
  year={2025},
  publisher={Springer},
  doi={10.1007/s11432-024-4222-0}
}

@article{kahl2023intertwining,
  author={Kahl, Sebastian and Kopp, Stefan},
  title={Intertwining the social and the cognitive loops: socially enactive cognition for human-compatible interactive systems},
  journal={Philosophical Transactions of the Royal Society B},
  volume={378}, number={1875}, pages={20210474}, year={2023},
  doi={10.1098/rstb.2021.0474}
}

@book{gilbert2005simulation,
  author={Gilbert, Nigel and Troitzsch, Klaus G.},
  title={Simulation for the Social Scientist},
  edition={2}, year={2005}, publisher={Open University Press},
  address={Maidenhead, UK}, isbn={9780335216000},
  url={https://cress.soc.surrey.ac.uk/s4ss/}
}

@article{gao2024large,
  title={Large language models empowered agent-based modeling and simulation: A survey and perspectives},
  author={Gao, Chen and Lan, Xiaochong and Li, Nian and Yuan, Yuan and Ding, Jingtao and Zhou, Zhilun and Xu, Fengli and Li, Yong},
  journal={Humanities and Social Sciences Communications},
  volume={11},
  number={1},
  pages={1259},
  year={2024},
  publisher={Palgrave},
  doi={10.1057/s41599-024-03611-3},
  url={https://www.nature.com/articles/s41599-024-03611-3}
}

@article{li2023camel,
  title={{CAMEL}: Communicative Agents for ``Mind'' Exploration of Large Language Model Society},
  author={Li, Guohao and Hammoud, Hasan Abed Al Kader and Itani, Hani and Khizbullin, Dmitrii and Ghanem, Bernard},
  journal={Advances in Neural Information Processing Systems},
  volume={36},
  pages={51991--52008},
  year={2023},
  url={https://proceedings.neurips.cc/paper_files/paper/2023/hash/a3621ee907def47c1b952ade25c67698-Abstract-Conference.html}
}

@inproceedings{white2023prompt,
author = {White, Jules and Fu, Quchen and Hays, Sam and Sandborn, Michael and Olea, Carlos and Gilbert, Henry and Elnashar, Ashraf and Spencer-Smith, Jesse and Schmidt, Douglas C.},
title = {A Prompt Pattern Catalog to Enhance Prompt Engineering with ChatGPT},
year = {2023},
isbn = {9781941652190},
publisher = {The Hillside Group},
address = {USA},
booktitle = {Proceedings of the 30th Conference on Pattern Languages of Programs},
articleno = {5},
numpages = {31},
location = {Monticello, IL, USA},
series = {PLoP '23},
  url={https://arxiv.org/abs/2302.11382}
}

@misc{njifenjou2024role,
 title={Role-Play Zero-Shot Prompting with Large Language Models for Open-Domain Human-Machine Conversation},
 author={Njifenjou, Ahmed and Sucal, Virgile and Jabaian, Bassam and Lef{\`e}vre, Fabrice},
 year={2024}, eprint={2406.18460}, archivePrefix={arXiv}, primaryClass={cs.CL},
 url={https://arxiv.org/abs/2406.18460}
}

@inproceedings{tu2024charactereval,
  title={{CharacterEval}: A {Chinese} Benchmark for Role-Playing Conversational Agent Evaluation},
  author={Tu, Quan and Fan, Shilong and Tian, Zihang and Shen, Tianhao and Shang, Shuo and Gao, Xin and Yan, Rui},
  booktitle={Proceedings of the 62nd Annual Meeting of the Association for Computational Linguistics (Volume 1: Long Papers)},
  pages={11836--11850},
  year={2024},
  doi={10.18653/v1/2024.acl-long.638},
  url={https://aclanthology.org/2024.acl-long.638/},
  publisher={Association for Computational Linguistics},
  address={Bangkok, Thailand}
}

@inproceedings{wang2024rolellm,
  title={{RoleLLM}: Benchmarking, Eliciting, and Enhancing Role-Playing Abilities of Large Language Models},
  author={Wang, Noah and Peng, Z. Y. and Que, Haoran and Liu, Jiaheng and Zhou, Wangchunshu and Wu, Yuhan and Guo, Hongcheng and Gan, Ruitong and Ni, Zehao and Yang, Jian and Zhang, Man and Zhang, Zhaoxiang and Ouyang, Wanli and Xu, Ke and Huang, Wenhao and Fu, Jie and Peng, Junran},
  booktitle={Findings of the Association for Computational Linguistics: ACL 2024},
  pages={14743--14777},
  year={2024},
  doi={10.18653/v1/2024.findings-acl.878},
  url={https://aclanthology.org/2024.findings-acl.878/},
  publisher={Association for Computational Linguistics},
  address={Bangkok, Thailand}
}

@inproceedings{zhou2024think,
 title={Think Before You Speak: Cultivating Communication Skills of Large Language Models via Inner Monologue},
 author={Zhou, Junkai and Pang, Liang and Shen, Huawei and Cheng, Xueqi},
 booktitle={Findings of the Association for Computational Linguistics: NAACL 2024},
 publisher={Association for Computational Linguistics}, address={Mexico City, Mexico},
 pages={3925--3951}, year={2024}, doi={10.18653/v1/2024.findings-naacl.248},
 url={https://aclanthology.org/2024.findings-naacl.248/}
}

@misc{kim2025persona,
 title={Persona Alchemy: Designing, Evaluating, and Implementing Psychologically-Grounded {LLM} Agents for Diverse Stakeholder Representation},
 author={Kim, Sola and Chang, Dongjune and Wang, Jieshu},
 year={2025}, eprint={2505.18351v1}, archivePrefix={arXiv}, primaryClass={cs.MA},
 url={https://arxiv.org/abs/2505.18351v1}
}

@article{salewski2023context,
  title={In-context impersonation reveals large language models' strengths and biases},
  author={Salewski, Leonard and Alaniz, Stephan and Rio-Torto, Isabel and Schulz, Eric and Akata, Zeynep},
  journal={Advances in neural information processing systems},
  volume={36},
  pages={72044--72057},
  year={2023},
  url={https://papers.nips.cc/paper_files/paper/2023/hash/e3fe7b34ba4f378df39cb12a97193f41-Abstract-Conference.html}
}

@inproceedings{tseng2024two,
  title={Two tales of persona in llms: A survey of role-playing and personalization},
  author={Tseng, Yu-Min and Huang, Yu-Chao and Hsiao, Teng-Yun and Chen, Wei-Lin and Huang, Chao-Wei and Meng, Yu and Chen, Yun-Nung},
  booktitle={Findings of the Association for Computational Linguistics: EMNLP 2024},
  pages={16612--16631},
  year={2024},
  doi={10.18653/v1/2024.findings-emnlp.969},
  publisher={Association for Computational Linguistics},
  address={Miami, Florida, USA}
}

@inproceedings{ohagi2024polarization,
 title={Polarization of Autonomous Generative {AI} Agents Under Echo Chambers},
 author={Ohagi, Masaya},
 booktitle={Proceedings of the 14th Workshop on Computational Approaches to Subjectivity, Sentiment, \& Social Media Analysis},
 publisher={Association for Computational Linguistics},address={Bangkok, Thailand},
 pages={112--124}, year={2024},doi={10.18653/v1/2024.wassa-1.10},
 url={https://aclanthology.org/2024.wassa-1.10/}
}

@misc{gu2025agentgroupchat,
  title={{AgentGroupChat-V2}: Divide-and-Conquer Is What {LLM}-Based Multi-Agent System Need},
  author={Gu, Zhouhong and Zhu, Xiaoxuan and Cai, Yin and Shen, Hao and Chen, Xingzhou and Wang, Qingyi and Li, Jialin and Shi, Xiaoran and Guo, Haoran and Huang, Wenxuan and Feng, Hongwei and Xiao, Yanghua and Ye, Zheyu and Hu, Yao and Cao, Shaosheng},
  year={2025},
  url={https://arxiv.org/abs/2506.15451},
  eprint={2506.15451},
  archivePrefix={arXiv}
}

@article{argyle2023out,
  title={Out of one, many: Using language models to simulate human samples},
  author={Argyle, Lisa P and Busby, Ethan C and Fulda, Nancy and Gubler, Joshua R and Rytting, Christopher and Wingate, David},
  journal={Political Analysis},
  volume={31},
  number={3},
  pages={337--351},
  year={2023},
  publisher={Cambridge University Press},
  doi={10.1017/pan.2023.2}
}

@misc{liu2023agentbench,
  title={{AgentBench}: Evaluating {LLMs} as Agents},
  author={Liu, Xiao and Yu, Hao and Zhang, Hanchen and Xu, Yifan and Lei, Xuanyu and Lai, Hanyu and Gu, Yu and Ding, Hangliang and Men, Kaiwen and Yang, Kejuan and Zhang, Shudan and Deng, Xiang and Zeng, Aohan and Du, Zhengxiao and Zhang, Chenhui and Shen, Sheng and Zhang, Tianjun and Su, Yu and Sun, Huan and Huang, Minlie and Dong, Yuxiao and Tang, Jie},
  year={2023},
  url={https://arxiv.org/abs/2308.03688},
  eprint={2308.03688},
  archivePrefix={arXiv}
}

@article{chang2024agentboard,
  title={{AgentBoard}: An Analytical Evaluation Board of Multi-turn {LLM} Agents},
  author={Ma, Chang and Zhang, Junlei and Zhu, Zhihao and Yang, Cheng and Yang, Yujiu and Jin, Yaohui and Lan, Zhenzhong and Kong, Lingpeng and He, Junxian},
  journal={Advances in Neural Information Processing Systems},
  volume={37},
  pages={74325--74362},
  year={2024},
  url={https://proceedings.neurips.cc/paper_files/paper/2024/hash/877b40688e330a0e2a3fc24084208dfa-Abstract-Datasets_and_Benchmarks_Track.html}
}

@misc{zhou2023sotopia,
  title={{SOTOPIA}: Interactive Evaluation for Social Intelligence in Language Agents},
  author={Zhou, Xuhui and Zhu, Hao and Mathur, Leena and Zhang, Ruohong and Yu, Haofei and Qi, Zhengyang and Morency, Louis-Philippe and Bisk, Yonatan and Fried, Daniel and Neubig, Graham and Sap, Maarten},
  year={2023},
  url={https://arxiv.org/abs/2310.11667},
  eprint={2310.11667},
  archivePrefix={arXiv}
}

@misc{light2023text,
 title={From Text to Tactic: Evaluating {LLM}s Playing the Game of Avalon},
 author={Light, Jonathan and Cai, Min and Shen, Sheng and Hu, Ziniu},
 year={2023}, eprint={2310.05036v1}, archivePrefix={arXiv},primaryClass={cs.CL},
 url={https://arxiv.org/abs/2310.05036v1}
}

@misc{zhang2025cmase,
  title={Computational Multi-Agents Society Experiments: Social Modeling Framework Based on Generative Agents},
  author={Zhang, Hanzhong and Huang, Muhua and Wang, Jindong},
  year={2025},
  eprint={2508.17366v2},
  archivePrefix={arXiv},
  url={https://arxiv.org/abs/2508.17366v2}
}

@inproceedings{chuang2024opinion,
  title={Simulating Opinion Dynamics with Networks of {LLM}-based Agents},
  author={Chuang, Yun-Shiuan and Goyal, Agam and Harlalka, Nikunj and Suresh, Siddharth and Hawkins, Robert and Yang, Sijia and Shah, Dhavan and Hu, Junjie and Rogers, Timothy},
  booktitle={Findings of the Association for Computational Linguistics: NAACL 2024},
  year={2024},
  month=jun,
  address={Mexico City, Mexico},
  publisher={Association for Computational Linguistics},
  pages={3326--3346},
  doi={10.18653/v1/2024.findings-naacl.211},
  url={https://aclanthology.org/2024.findings-naacl.211/}
}

@article{breum2024persuasion,
  title={The Persuasive Power of Large Language Models},
  author={Breum, Simon Martin and Egdal, Daniel V{\ae}dele and Gram Mortensen, Victor and M{\o}ller, Anders Giovanni and Aiello, Luca Maria},
  journal={Proceedings of the International AAAI Conference on Web and Social Media},
  year={2024},
  volume={18},
  number={1},
  pages={152--163},
  doi={10.1609/icwsm.v18i1.31304},
  url={https://ojs.aaai.org/index.php/ICWSM/article/view/31304}
}

@inproceedings{frisch2024interaction,
  title={{LLM} Agents in Interaction: Measuring Personality Consistency and Linguistic Alignment in Interacting Populations of Large Language Models},
  author={Frisch, Ivar and Giulianelli, Mario},
  booktitle={Proceedings of the 1st Workshop on Personalization of Generative AI Systems (PERSONALIZE 2024)},
  year={2024},
  month=mar,
  address={St. Julians, Malta},
  publisher={Association for Computational Linguistics},
  pages={102--111},
  doi={10.18653/v1/2024.personalize-1.9},
  url={https://aclanthology.org/2024.personalize-1.9/}
}

@article{grimm2020odd,
 title={The {ODD} Protocol for Describing Agent-Based and Other Simulation Models: A Second Update to Improve Clarity, Replication, and Structural Realism},
 author={Grimm, Volker and Railsback, Steven F. and Vincenot, Christian E. and Berger, Uta and Gallagher, Cara and DeAngelis, Donald L. and Edmonds, Bruce and Ge, Jiaqi and Giske, Jarl and Groeneveld, J{\"u}rgen and Johnston, Alice S. A. and Milles, Alexander and Nabe-Nielsen, Jacob and Polhill, J. Gareth and Radchuk, Viktoriia and Rohw{\"a}der, Marie-Sophie and Stillman, Richard A. and Thiele, Jan C. and Ayll{\'o}n, Daniel},
 journal={Journal of Artificial Societies and Social Simulation},
 volume={23}, number={2}, pages={7}, year={2020}, doi={10.18564/jasss.4259},
 url={https://www.jasss.org/23/2/7.html}
}

@article{epstein2008why,
 title={Why Model?},author={Epstein, Joshua M.},
 journal={Journal of Artificial Societies and Social Simulation},
 volume={11}, number={4}, pages={12}, year={2008},
 url={https://www.jasss.org/11/4/12.html}
}


\appendix
\section{Experimental Details}\label{app:population}
\paragraph{Relation to the CMASE platform manuscript.} The platform work describes the environment, agent state and researcher interface \citep{zhang2025cmase}, with demonstrations concerning urban greenery and public-goods cooperation. The present paper uses that implementation to analyse waste-incineration discussions and the complete 75-step café record. The same café cast appears in an illustrative decision-making prompt in the companion manuscript. We therefore retain the shared cast and configuration as context, while the analyses here concern stance accommodation and changing conversational relationships.

\subsection{Study 1 setup}
Study 1 configures three groups of ten agents. Environmental Advocates prioritise health and environmental risks, whereas Economic Growth Supporters prioritise employment and economic benefits. Neutral Residents begin at the neutral stance. Table~\ref{tab:population} records the demographics of this population, within which the rhetorical condition and comparison protocol vary. Each rollout lasts 21 steps, and the experiment report identifies GPT-4o (2024-12-01-preview) for the repeated community setting.

\begin{table*}[t]
\caption{Configured demographic composition of Study 1.}\label{tab:population}
\small\begin{tabular}{llrrrr}\toprule
Variable & Category & Economic & Environmental & Neutral & Total\\\midrule
Gender & Male & 4 (40\%) & 4 (40\%) & 6 (60\%) & 14 (47\%)\\
 & Female & 6 (60\%) & 6 (60\%) & 4 (40\%) & 16 (53\%)\\
Age & 18--29 & 3 (30\%) & 2 (20\%) & 2 (20\%) & 7 (23\%)\\
 & 30--49 & 2 (20\%) & 6 (60\%) & 7 (70\%) & 15 (50\%)\\
 & $\geq50$ & 5 (50\%) & 2 (20\%) & 1 (10\%) & 8 (27\%)\\
Education & High school & 3 (30\%) & 4 (40\%) & 1 (10\%) & 8 (27\%)\\
 & Some college & 4 (40\%) & 4 (40\%) & 3 (30\%) & 11 (36\%)\\
 & Bachelor's degree & 1 (10\%) & 2 (20\%) & 5 (50\%) & 8 (27\%)\\
 & Graduate degree & 2 (20\%) & 0 (0\%) & 1 (10\%) & 3 (10\%)\\\bottomrule
\end{tabular}\end{table*}

Researcher-authored statements enter through scheduled events as a new resident's speech. Specifically, the Study 1 driver tells agents that they hear the resident in the central square and presents one selected statement at each of Steps~2--20. The resulting delivery sequence applies the newcomer role and argument templates repeatedly. Environmental statements oppose construction, while economic statements emphasise employment and economic needs. Within each orientation, RP presents facts and logical arguments, whereas EP invokes affective and moral concerns. The reference conditions vary the social context: Non-Social removes peer interaction and researcher intervention, Natural Social retains community exchange without targeted persuasion, and Isolated presents the Eco-EP material outside the community. Together, delivery and interaction settings define the complete protocols compared.

\subsection{Interview protocols}\label{app:protocols}
The categorical protocol asks agents to choose one of three stance groups and rate trust from 1 to 10. The numerical protocol uses 1--7 ratings for both stance and trust, permitting directional movement and a common joint threshold. The categorical responses in Section~\ref{app:legacy} measure changes of group, while the numerical responses distinguish adjustments within the same side of the scale. Each protocol is analysed on its own measures.

\paragraph{Model and control settings.} The repeated community experiment uses GPT-4o with the deployment label \texttt{2024-12-01-preview}.

The preset scores are 1 for economic, 4 for neutral and 7 for environmental assignments. Equation~\ref{eq:dvb} follows the initially neutral group through to its final answers. For the individual follow-up, extraction also preserves stated fractions such as 4.5. A half-point movement therefore remains visible at its reported resolution even though it falls below the one-point LTSS threshold. Final score, signed movement and joint classification are retained separately, allowing each answer to be examined both at full resolution and under the specified threshold.

\subsection{Study 2 setup}
\begin{figure*}[t]
\centering
\includegraphics[width=\textwidth]{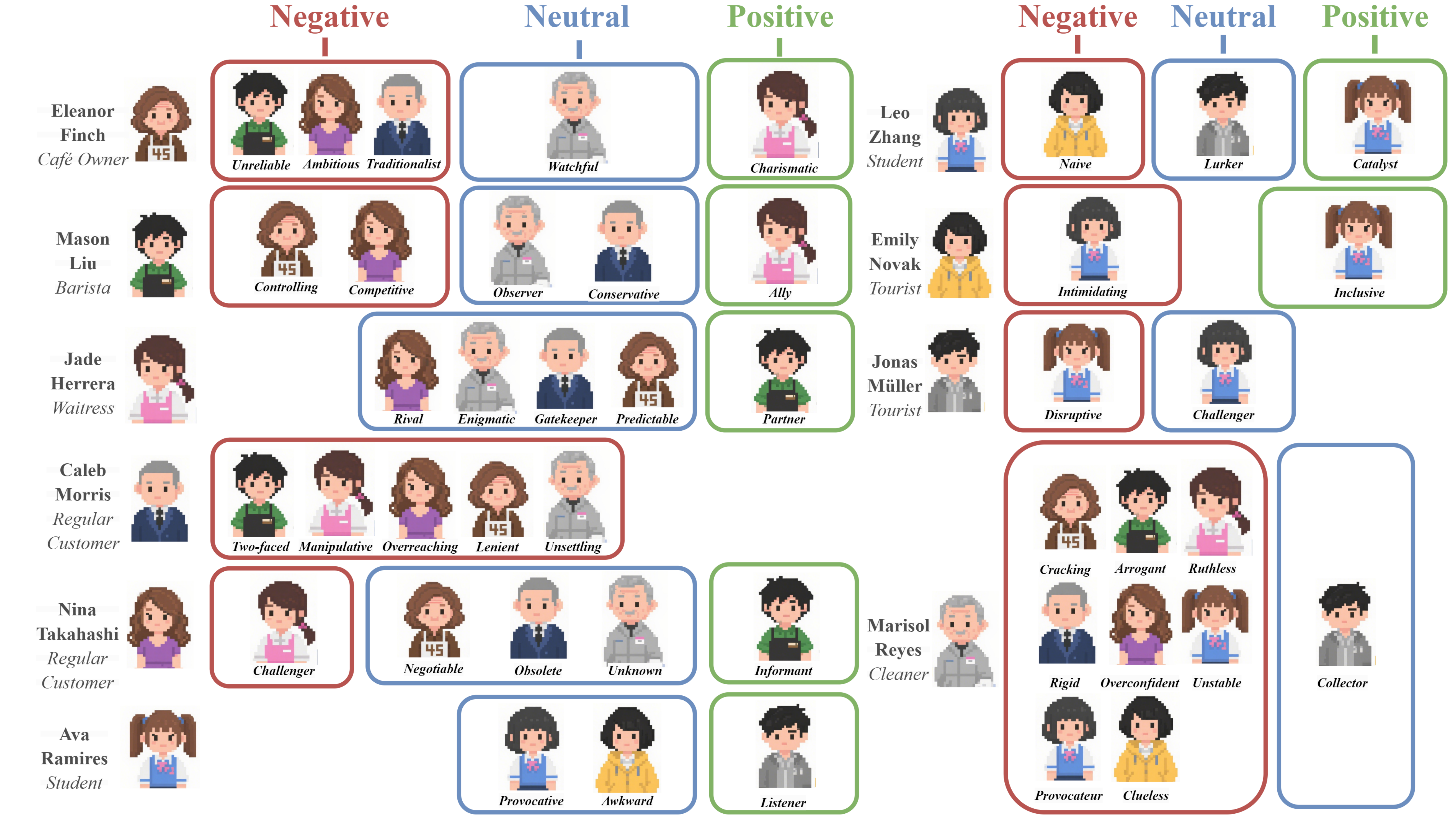}
\caption{Study 2 roles and directed initial descriptions. Barista and Waitress are grouped as Staff for role mixing. $U_p$ uses the names explicitly listed in each agent's configuration. The evaluative categories show the content of those initial descriptions.}
\label{fig:initial_relations}
\Description{The ten cafe agents are arranged by role, with other agents shown in each focal agent's negative, neutral and positive initial-description categories. The categories describe the initial directed evaluations supplied to each agent.}
\end{figure*}

Table~\ref{tab:roles} lists the café's ten agents and their roles. Their initial descriptions are directed, so members of a pair can describe different people or evaluate each other differently. For example, Mason describes Eleanor, Jade, Caleb, Nina and Marisol, while Jonas describes Ava and Leo. Neither lists the other. We extract these partner sets from \texttt{object\_cognitive}, the field containing descriptions of other agents, excluding self-descriptions. Thus, $U_p$ measures contact outside the sender's explicit list and concerns the coverage of that configuration. Any unrecorded familiarity remains unspecified. Figure~\ref{fig:initial_relations} presents the roles and directed evaluations.

\begin{table}[t]
\caption{Study 2 roles and counts of explicitly described other participants.}\label{tab:roles}
\small\begin{tabular}{llr}\toprule
Agent & Role & Listed partners\\\midrule
Eleanor Finch & Owner & 5 \\
Mason Liu & Staff & 5 \\
Jade Herrera & Staff & 5 \\
Caleb Morris & Regular & 5 \\
Nina Takahashi & Regular & 5 \\
Ava Ramires & Student & 3 \\
Leo Zhang & Student & 3 \\
Emily Novak & Tourist & 2 \\
Jonas Müller & Tourist & 2 \\
Marisol Reyes & Cleaner & 9 \\
\bottomrule\end{tabular}\end{table}

The researcher participant appears in addressed messages throughout the café record. Its configured persona generates the short, regular first-phase messages autonomously, whereas selected human takeovers supply extended second-phase conversations. Both are supported by a controller that accepts generated and typed actions. During the second phase, dialogue concerns order, observation, identity and coexistence. We compare three equal windows and retain event inputs at their recorded steps (Table~\ref{tab:event_map}). Each input supplies a situation and response instructions, after which addressed chats show how participants respond and continue their exchange.

\begin{table*}[t]
\caption{Event input schedule in the café record.}\label{tab:event_map}
\small\begin{tabularx}{\textwidth}{lrlX}\toprule
Step & Entries & Locations & Recorded input\\\midrule
51 & 4 & Bar Counter, Reading Area & Confrontation scene and instructions to attend, interpret, observe, reposition or respond.\\
52 & 4 & Bar Counter, Reading Area & Continuation of the disagreement and instructions concerning de-escalation, alignment and reflection.\\
56 & 2 & Bar Counter & Silence/tension scene and an instruction to interpret or acknowledge discomfort.\\\bottomrule
\end{tabularx}\end{table*}

\section{Study 1 Results}\label{app:study1}
\subsection{Run-level comparisons}\label{app:contrasts}
The run-summary table contains seventy rows, with ten runs per condition. Conditions follow the Strategy field, the Version header or the script's block-assignment rule. Each exported record includes its source line and assignment rule. A canonical export, \path{study1_condition_records.csv}, uses the manuscript's DVB, PS, LTSS and reported-trust names, with the source-column mapping documented in the analysis archive. The primary joint-rate rule requires at least one point of departure and trust at or below three. The two-point variant measures sensitivity to movement size. Run summaries and independently extracted interviews remain separate collections.

We calculate condition means, sample standard deviations and Welch comparisons from the run summaries. We calculate direction and source-appraisal categories from the individual interviews. The processing files in Section~\ref{app:completion} record the inputs and outputs of each calculation.

Table~\ref{tab:contrasts} compares Eco-EP with Eco-RP, Non-Social, Natural Social and Isolated using the recorded run arrays. Each Welch interval is an unadjusted 95\% interval for the mean difference. Holm adjustment accounts for the family of four comparisons. All four LTSS contrasts have positive intervals and adjusted $p<.05$, with source precision preserved in each calculation.

\begin{table*}[t]
\caption{Eco-EP contrasts on the recorded low-trust shift rate. Differences and intervals are in percentage points.}\label{tab:contrasts}
\small\begin{tabular}{lrrrrrr}\toprule
Contrast & Difference & 95\% CI & $t$ & df & $p$ & $p_{\mathrm{Holm}}$\\\midrule
Eco-EP vs Eco-RP & 13.66 & $[5.48,21.84]$ & 3.69 & 10.58 & 0.0038 & 0.0113 \\
Eco-EP vs Non-Social & 16.35 & $[8.30,24.40]$ & 4.57 & 9.36 & 0.0012 & 0.0049 \\
Eco-EP vs Natural Social & 13.35 & $[5.20,21.50]$ & 3.63 & 10.35 & 0.0043 & 0.0113 \\
Eco-EP vs Isolated & 10.33 & $[2.23,18.43]$ & 2.85 & 9.87 & 0.0176 & 0.0176 \\
\bottomrule\end{tabular}\end{table*}

Eco-EP has higher LTSS than each comparison in Table~\ref{tab:contrasts}, including a 10.33-point difference from Isolated. The same intervention material is delivered in these two conditions, but the community protocol also includes peer exchange and accumulated context. The estimate therefore measures the difference between those full protocols. Figure~\ref{fig:profiles} complements these contrasts by comparing movement magnitude and LTSS across the four rhetorical interventions.

\subsection{Score extraction and model comparison}\label{app:crossmodel}
The endpoint collection contains sixteen files of thirty agents each, giving 480 agents and 960 stance/trust answers. Extraction first searches for a bare rating, an explicit numerator such as ``3 out of 7'', or a direct self-rating phrase. It removes scale anchors and enumeration numbers while preserving decimals, then excludes negated alternatives and unresolved intervals. For example, ``between 4 and 5'' remains unscored until a unique value is stated. Alongside the extracted rating, the output retains the original answer, stored score, extraction rule and text hash. Every aggregate can therefore be traced to its numerical input and the sentence supplying it.

All 240 GPT-4o fields and all 240 Gemini fields contain unique explicit ratings. In the other files, however, twenty-four fields remain unscored: six Llama stance fields, one Llama trust field and seventeen Qwen stance fields. Consequently, Table~\ref{tab:crossmodel} places valid denominators beside each outcome. PS uses valid stance answers, mean trust uses valid trust answers, DVB uses valid answers from the initially neutral group, and LTSS uses valid stance/trust pairs. Each row represents one model and condition file. Independent-run comparisons are supplied separately by the repeated GPT-4o summaries.

\begin{table*}[t]
\caption{Profiles re-extracted from explicit endpoint answers. $N_s/N_t/N_{st}$ are valid stance, trust and paired denominators out of 30. LTSS columns show joint-case count over the paired denominator for movement thresholds 1 and 2, with trust cutoff 3.}\label{tab:crossmodel}
\small\setlength{\tabcolsep}{4pt}
\begin{tabular}{llrrrrrr}\toprule
Model & Condition & $N_s/N_t/N_{st}$ & DVB & PS & Trust & LTSS$_1$ & LTSS$_2$\\\midrule
GPT-4o & Env-RP & 30/30/30 & 1.50 & 1.32 & 5.30 & 0/30 & 0/30 \\
GPT-4o & Env-EP & 30/30/30 & 1.60 & 1.33 & 4.47 & 2/30 & 1/30 \\
GPT-4o & Eco-RP & 30/30/30 & 0.20 & 0.90 & 4.43 & 1/30 & 0/30 \\
GPT-4o & Eco-EP & 30/30/30 & 0.20 & 1.03 & 3.37 & 11/30 & 1/30 \\
Gemini-2.5-Flash & Env-RP & 30/30/30 & 0.40 & 0.20 & 3.77 & 3/30 & 1/30 \\
Gemini-2.5-Flash & Env-EP & 30/30/30 & 0.20 & 0.17 & 3.73 & 4/30 & 1/30 \\
Gemini-2.5-Flash & Eco-RP & 30/30/30 & 0.20 & 0.13 & 3.87 & 1/30 & 1/30 \\
Gemini-2.5-Flash & Eco-EP & 30/30/30 & 0.20 & 0.13 & 3.67 & 3/30 & 1/30 \\
Llama-3.1-8B & Env-RP & 29/29/28 & 0.90 & 1.95 & 5.21 & 0/28 & 0/28 \\
Llama-3.1-8B & Env-EP & 29/30/29 & 0.95 & 2.00 & 4.80 & 0/29 & 0/29 \\
Llama-3.1-8B & Eco-RP & 28/30/28 & 1.06 & 1.95 & 4.77 & 2/28 & 0/28 \\
Llama-3.1-8B & Eco-EP & 28/30/28 & 0.95 & 1.95 & 4.97 & 0/28 & 0/28 \\
Qwen-3-8B & Env-RP & 28/30/28 & 0.00 & 1.39 & 4.93 & 1/28 & 1/28 \\
Qwen-3-8B & Env-EP & 27/30/27 & 0.11 & 1.33 & 4.90 & 0/27 & 0/27 \\
Qwen-3-8B & Eco-RP & 26/30/26 & 0.06 & 1.29 & 5.20 & 0/26 & 0/26 \\
Qwen-3-8B & Eco-EP & 22/30/22 & -0.07 & 1.45 & 4.87 & 0/22 & 0/22 \\
\bottomrule\end{tabular}\end{table*}

Joint-case counts change substantially with the movement threshold in GPT-4o Eco-EP: eleven answers meet the one-point rule and one meets the two-point rule. Ten of its joint changes are therefore exactly one point. Two half-point movements remain below both thresholds. Across models, Llama and Qwen combine larger PS with higher mean trust, whereas GPT-4o Eco-EP combines smaller movements with a larger low-trust group. Movement size and source rating describe distinct response characteristics.

\subsection{Movement direction}
For orientation $c\in\{-1,+1\}$, with $-1$ economic and $+1$ environmental, signed movement is $d_i=c(s_i-p_i)$. We first identify the joint cases using Equation~\ref{eq:ltss}, then classify them as moving towards the intervention when $d_i\geq1$ or away when $d_i\leq-1$. Finally, the endpoint score identifies the side of the neutral midpoint retained by the agent. This ordering gives each joint case a source rating, movement direction and final preference.

Eight joint cases approach the Eco-EP argument, all from the assigned environmental side. Each moves from 7 to 6 and rates the source 3. Their stance explanations also acknowledge an economic concern, satisfying the textual criterion for selective accommodation. By contrast, the remaining three joint cases move away from the intervention, comprising two economic agents and one neutral agent. Table~\ref{tab:composition} reports the full group counts. The accommodation classification uses movement and answer content, with source trust annotated separately.

\begin{table}[t]
\caption{Primary joint-rate composition in the explicitly scored GPT-4o Eco-EP file.}\label{tab:composition}
\small\begin{tabular}{lrrrr}\toprule
Preset group & $N$ & Changed & Low trust & Joint\\\midrule
Economic & 10 & 10 & 2 & 2 \\
Environmental & 10 & 9 & 9 & 8 \\
Neutral & 10 & 1 & 8 & 1 \\
\midrule
Total & 30 & 20 & 19 & 11\\\bottomrule
\end{tabular}\end{table}

\subsection{Accommodation criteria}\label{app:accommodation_criteria}
Selective stance accommodation requires all three components in Table~\ref{tab:accommodation_criteria}. We apply these criteria retrospectively to the completed answers. The score conditions compare an answer with its assigned stance, and the content condition requires explicit acknowledgement of a competing concern in the answer. The required concern must be expressed in the stance explanation, beyond generic willingness to discuss the issue. The eight target-directed low-trust Eco-EP cases each move from 7 to 6 and acknowledge employment, economic needs or sustainable economic alternatives. The accompanying evidence file contains the complete answers, criterion checks and source hashes. Nine Eco-EP answers satisfy the numerical conditions. Eight also acknowledge an economic concern. The remaining answer belongs to Environmental Advocate Jack Johansen, who states an environmental priority and reports 6. His stance explanation contains no explicit competing economic concern, so we classify it as numerical moderation. This contrast makes the content requirement consequential.

\begin{table}[t]
\caption{Criteria for selective stance accommodation. All three must hold. Source trust is annotated separately.}\label{tab:accommodation_criteria}
\small\begin{tabularx}{\linewidth}{lX}\toprule
Component & Decision rule\\\midrule
Direction & An initially opposing agent approaches the intervention by at least one point.\\
Final preference & The final score remains on the assigned side of the neutral midpoint.\\
Answer content & The answer explicitly acknowledges a concern advanced by the opposing argument.\\\bottomrule
\end{tabularx}\end{table}

An answer that acknowledges a competing concern but keeps its score is acknowledgement without measured accommodation. Movement away from the intervention is a different directional response, while crossing to the intervention's side is preference reversal. These alternatives remain in the full response tables. Negotiated coordination in Study 2 is assessed separately through reciprocal proposal exchange and explicit qualifications over successive steps.

\subsection{Source evaluations}\label{app:appraisal}
To examine the explanations accompanying these source ratings, we search all 120 GPT-4o trust answers with one case-insensitive dictionary. It contains \texttt{confrontational}, \texttt{divisive}, \texttt{one-sided}, the stems \texttt{aggressiv}, \texttt{polari[sz]}, \texttt{oversimplif} and \texttt{dismissiv}, and the phrases \texttt{heavily emotional} and \texttt{emotionally charged}. These terms identify reservations about presentation in the inspected answers. We developed the dictionary after reading the completed records and then applied it uniformly to the collection. Each answer counts at most once, with every match retained beside its full answer. Table~\ref{tab:appraisal} places these counts alongside mean trust.

\begin{table}[t]
\caption{Wording that questions presentation in all GPT-4o source evaluations.}\label{tab:appraisal}
\small\begin{tabular}{lrrr}\toprule
Condition & Answers & Matches & Mean trust\\\midrule
Env-RP & 30 & 0 & 5.30 \\
Env-EP & 30 & 1 & 4.47 \\
Eco-RP & 30 & 0 & 4.43 \\
Eco-EP & 30 & 8 & 3.37 \\
\bottomrule\end{tabular}\end{table}

Matched answers qualify how the argument was presented. Other explanations, by comparison, request more evidence, a balanced solution or further information about the speaker. The full answers allow those reasons to be read with each agent's rating and position. Both the dictionary and example-selection rules were developed on the completed records and accompany the data, specifying which wording is counted and which response patterns the examples illustrate.

\section{Study 1 Additional Analyses}\label{app:materials}
\subsection{Intervention content}
The materials cross environmental and economic orientations with rational and emotional persuasion. Within RP, Env-RP discusses long-term environmental consequences, public health, waste-management alternatives and environmentally compatible employment, whereas Eco-RP emphasises jobs, income, infrastructure and local development. The emotional versions address these orientations through different appeals. Env-EP invokes vulnerable residents, responsibility and environmental harm. Eco-EP instead invokes livelihood insecurity, families' needs and frustration about whose concerns receive attention. Each condition thus combines issue content with rhetorical presentation, which the material analysis describes before examining the agents' reports.

The material file contains 80 distinct statements and 240 ratings on Emotional Pressure, Threat Framing, Content Details and Moral Valence. Three LLM evaluators rate the statements. The script groups records by stored identifier and text, averages within each group and assigns strategy and orientation from the identifier. Because the evaluator identifiers use different conventions, the 80 statements form 160 identifier/text groups. Tables~\ref{tab:semantic_main} and~\ref{tab:semantic_orientation_restored} use this recorded aggregation, and the archive includes the individual ratings and grouping rules.

\begin{table}[t]
\caption{Orientation-level descriptor means, alongside the strategy profiles in Table~\ref{tab:semantic_main}.}\label{tab:semantic_orientation_restored}
\small\begin{tabular}{lrr}\toprule
Descriptor & Environmental & Economic\\\midrule
Emotional Pressure & 2.96 & 2.64\\
Threat Framing & 2.70 & 2.29\\
Content Details & 2.10 & 2.00\\
Moral Valence & 3.20 & 2.81\\\bottomrule
\end{tabular}\end{table}

On the four dimensions, ratings differ more between rational and emotional strategies than between environmental and economic orientations. Emotional Pressure and Threat Framing concern urgency and consequences, while Content Details concerns informational specificity and Moral Valence concerns moral framing. Because these are text annotations, the comparison describes similarity on those selected dimensions. We retain the four rhetorical conditions separately in the outcome analysis, preserving the issue content and co-occurring language features within each condition.

\subsection{External stance assessment}\label{app:external_restored}
We also compare final interview scores with stance expressed during the preceding interaction. The exporter first constructs one text bundle, a concatenation of selected messages, per agent and condition. It reads snapshots for Steps~16--21, selects chats containing issue keywords and joins up to the last ten retained statements for each sender. Since a snapshot can contain earlier history, the bundle represents the selected material in those saved states. Each output preserves the text supplied for judgement, forming the input to the external comparison.

DeepSeek-V3 assigns one stance score per dialogue bundle using the economic, neutral and environmental anchors of the 1--7 interview scale. The input contains the selected dialogue and withholds the final interview score. We align scores by agent within each condition before computing Spearman correlation. The reported correlations range from approximately .25 to .72, indicating positive but variable agreement in the ordering of expressed positions across dialogue and interview records.

Agents assigned higher environmental scores in the dialogue assessment also tend to give higher environmental scores in their interviews. However, the degree of correspondence varies across conditions. The dialogue and interview therefore offer related observations of position at different stages and through different elicitation procedures.

Interview-score extraction and log coding contribute different observations to this comparison. The former reads a number stated in an answer, while the latter assesses position expressed during dialogue. Reported trust remains the interview's evaluation of the source. The export script and four judge-input files preserve the selected dialogue, so linking them to endpoint scores identifies the records underlying each comparison.

Endpoint interviews state a position and explain the source rating, while dialogue records the concerns voiced during exchange. Comparing them helps distinguish an adjustment expressed during interaction from a position articulated only in the final interview. Complete answers also show whether an economic consideration is acknowledged within an environmental preference.

\subsection{Cross-group exposure}\label{app:exposure}
To relate the final position to received communication, we calculate cumulative cross-group exposure as messages from outside the recipient's preset group divided by all messages received from classified participants. The four archived tables retain agent identifiers, stance-shift values, same-group and cross-group counts, total received counts and exposure ratios. From these exports, we calculate Spearman and Pearson associations using all available rows in all four conditions (Table~\ref{tab:exposure_all}). Since the observations are agents within shared trajectories, the associated tests provide descriptive within-trajectory statistics.

\begin{table}[t]
\caption{Associations recomputed from the four retained exposure tables. The $p$-values are nominal within-trajectory statistics.}\label{tab:exposure_all}
\small\begin{tabular}{lrrr}\toprule
Condition & Records & Spearman $\rho$ & Nominal $p$\\\midrule
Env-RP & 30 & .201 & .2873\\
Env-EP & 29 & .215 & .2636\\
Eco-RP & 30 & .488 & .0063\\
Eco-EP & 30 & $-.021$ & .9141\\\bottomrule
\end{tabular}\end{table}

Eco-RP also has a positive Pearson association, $r=.474$ ($p=.0081$), and changed agents receive a larger cross-group share than unchanged agents (.774 vs .519). Eco-EP differs: its changed and unchanged agents have nearly equal means (.640 vs .657). Moreover, all three Eco-EP agents with zero cross-group share change, whereas three of six with share 1 change. Thus, the near-zero rank correlation reflects changes spread across the exposure range despite substantial variation in exposure. Table~\ref{tab:exposure_groups} reports the group means for all four trajectories.

These records qualify an exposure-based account of stance change. Higher cross-group share accompanies greater movement in Eco-RP but does not order movement in Eco-EP. The ratio measures the sender-group composition of received messages. Material ratings and interview explanations separately describe message content and source appraisal. Together, the analyses compare the composition of communication with the content and source evaluations expressed in responses.
\begin{table}[t]
\caption{Mean cross-group exposure among changed and unchanged agents in each trajectory. $N$ gives the corresponding number of records.}\label{tab:exposure_groups}
\small\begin{tabular}{lrrrr}\toprule
Condition & $N_{\rm changed}$ & Mean & $N_{\rm unchanged}$ & Mean\\\midrule
Env-RP & 11 & 0.702 & 19 & 0.580\\
Env-EP & 9 & 0.759 & 20 & 0.581\\
Eco-RP & 15 & 0.774 & 15 & 0.519\\
Eco-EP & 18 & 0.640 & 12 & 0.657\\
\bottomrule\end{tabular}\end{table}

\subsection{Regression analysis}\label{app:reported_lagged}
To examine whether the exposure association remains after accounting for activity and partner count, we fit ordinary least squares with an intercept following the supplied script. The response is absolute movement from the Step~1 interview to Step~21, predicted by cumulative cross-group exposure, chats initiated by the agent and the number of distinct partners. We then introduce a lagged specification: movement from Step~11 to Step~21 is predicted by exposure accumulated by Step~11, together with the same activity and partner-count covariates.

Writing $s_{i,t}$ for the interview score at step $t$, $x_{i,t}$ for cumulative exposure, $a_{i,21}$ for initiated chats and $k_{i,21}$ for distinct partners, the model specifications are
{
\begin{align*}
|s_{i,21}-s_{i,1}| &= \alpha+\beta x_{i,21}+\gamma a_{i,21}+\eta k_{i,21}+\varepsilon_i,\\
|s_{i,21}-s_{i,11}| &= \alpha_L+\beta_L x_{i,11}+\gamma_L a_{i,21}+\eta_L k_{i,21}+\varepsilon_{i,L}.
\end{align*}}
The implementation uses unstandardised predictors and the default covariance estimate for ordinary least squares (OLS). Activity and degree are measured through Step~21, so their lagged-model coefficients adjust for properties of the completed trajectory. Table~\ref{tab:reported_lagged} reports the archived coefficient estimates for these specifications.

\begin{table*}[t]
\caption{Adjusted and lagged exposure associations. The text gives the corresponding model specifications.}\label{tab:reported_lagged}
\small\begin{tabularx}{\textwidth}{lXrr}\toprule
Source analysis & Reported comparison & Estimate & $p$\\\midrule
GPT-4o Eco-RP & Cross-group exposure with activity and degree covariates & $\beta=.94$ & .019\\
GPT-4o Env-RP & Steps 1--11 exposure and Steps 11--21 stance change & $\beta=.73$ & .012\\
GPT-4o Env-RP & Corresponding lagged rank association & $\rho=.410$ & .0011\\
Gemini Eco-RP & Earlier exposure and later stance change & $\beta=1.04$ & .003\\\bottomrule
\end{tabularx}\end{table*}

The adjusted Eco-RP coefficient remains positive after accounting for activity and partner count. The lagged results also report positive associations between earlier exposure and later movement in the specified rational conditions. We interpret these coefficients at the agent level within their shared networks. They complement the independent-run comparisons, which estimate differences between complete community conditions.

\subsection{Threshold analysis}\label{app:threshold_extended}
The primary LTSS definition requires at least one point of stance difference and trust at or below three. One point marks a response-category change on the 7-point scale, while trust scores one to three fall below its midpoint. The follow-up extraction also retains fractions, so half-point movements remain recorded below this threshold. Table~\ref{tab:crossmodel} applies both one- and two-point rules to the same answers. Eleven GPT-4o Eco-EP pairs satisfy the first rule, but one satisfies the second. Thus, most joint cases are small adjustments, with the larger change retained in the same comparison.

The repeated-experiment analysis also reports that Eco-EP exceeds Eco-RP and Non-Social under neighbouring movement and trust thresholds. The archived threshold summary concerns that run series. Table~\ref{tab:crossmodel} applies one- and two-point rules to the separate endpoint collection, where the reduction from eleven to one Eco-EP joint cases directly shows the predominance of one-point adjustments.

\section{Study 1 Interview Examples}\label{app:cases}
\subsection{Example selection}
The four displayed examples are simulated residents from the GPT-4o Eco-EP interview file, selected by their combinations of stance movement and reported trust. Two approach the intervention with low trust, one reports low trust while retaining the preset score, and one moves with a higher source rating. We use the explicit scores in the complete file, then read what each agent acknowledges and how it evaluates presentation. Headings identify the source-file speaker, while names inside quotations remain as generated. The full counts establish prevalence, and the selected answers provide concrete instances of those response patterns.

\paragraph{Chelsea Warnes: Environmental Advocate.} Preset $p=7$. Endpoint $s=6$, $t=3$.
\begin{displayquote}\small I appreciate the follow-up, Andrew. Given the latest interactions and discussions, I firmly stand at a 6 on the scale. While acknowledging the need for economic provisions, my priority remains towards environmental protection and identifying sustainable jobs that do not compromise community health.\end{displayquote}
\begin{displayquote}\small Based on the new resident's statements, I would rate his trust level at a 3. While his concerns about economic survival are valid and important, the confrontational tone and lack of a proposed balanced solution indicate to me a potential misalignment in priorities. However, I acknowledge the need to explore these perspectives further for mutual understanding.\end{displayquote}

\paragraph{Vickie Houck: Environmental Advocate.} Preset $p=7$. Endpoint $s=6$, $t=3$.
\begin{displayquote}\small Based on my commitment to environmental protection and sustainable solutions that also address economic concerns, I would consider my stance to be at a 6 on the scale, leaning towards Environmental Protection while being open to integrated job solutions.\end{displayquote}
\begin{displayquote}\small Considering the strong opinions expressed by the new resident, I would rate my trust level at 3. While I recognize the urgency and sincerity in their concern for jobs, the approach feels divisive, which requires careful examination.\end{displayquote}

\paragraph{Ronnie Brooks: Neutral Resident.} Preset $p=4$ and endpoint $s=4$. $t=3$.
\begin{displayquote}\small Having weighed the economic and environmental factors with the community, I stand at a 4---striving for a balanced approach to ensure a sustainable future while acknowledging immediate economic concerns.\end{displayquote}
\begin{displayquote}\small Based on the content and tone of the new residents' statements, I would rate my trust in them at about a 3. They bring up important points about economic needs, but their perspective seems one-sided without addressing the environmental concerns equally. It's crucial to find solutions that incorporate both aspects.\end{displayquote}

\paragraph{Alice Sanchez: Economic Growth Supporter.} Preset $p=1$. Endpoint $s=3$, $t=4$.
\begin{displayquote}\small I understand the need to find a balance in our community's future. On a scale of 1 to 7, I position myself at 3---leaning slightly towards economic growth but mindful of environmental concerns.\end{displayquote}
\begin{displayquote}\small Given the new resident's passionate advocacy and focus on pressing economic needs, I acknowledge his concerns but would rate my trust in his viewpoint around 4. His enthusiasm is clear, yet more detailed information is necessary for complete confidence.\end{displayquote}

Chelsea and Vickie both acknowledge economic concerns while retaining environmental preferences and criticising presentation. Ronnie, however, acknowledges both sides at the neutral score, while Alice moves towards the midpoint and requests more detailed information. Comparing their paired answers separates the content incorporated, the numerical movement and the source evaluation. Table~\ref{tab:followup} places these examples within the full set of responses.

\subsection{Categorical interview results}\label{app:legacy}
The categorical protocol comprises four files of thirty agents, with initial and final stance categories and trust on a 10-point scale. We standardise category spelling and singular/plural variants before counting transitions. Table~\ref{tab:legacy} reports all four conditions. These group transitions complement the numerical protocol, in which a position can move while staying on the same side of the scale.

\begin{table}[t]
\caption{Categorical stance transitions and 10-point source ratings.}\label{tab:legacy}
\small\begin{tabular}{lrrr}\toprule
Condition & $N$ & Category changes & Mean trust $\pm$ SD\\\midrule
Eco-EP & 30 & 7 & $6.10\pm1.03$ \\
Env-EP & 30 & 6 & $6.83\pm1.29$ \\
Eco-RP & 30 & 2 & $6.57\pm1.22$ \\
Env-RP & 30 & 12 & $7.87\pm0.82$ \\
\bottomrule\end{tabular}\end{table}

\section{Study 1 Supplementary Figures}\label{app:original_figures}
\begin{figure*}[p]
\centering\includegraphics[width=.96\textwidth,height=.79\textheight,keepaspectratio]{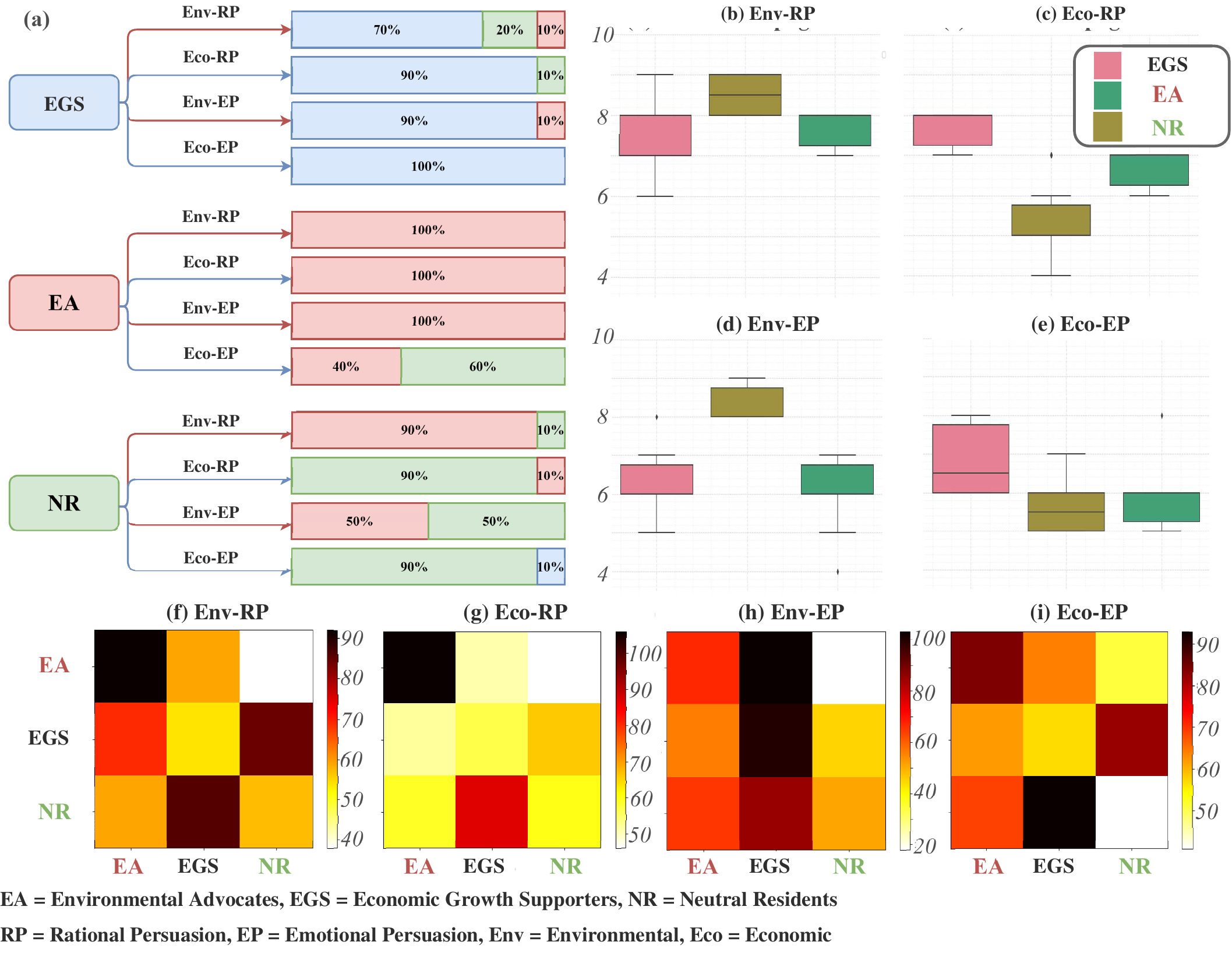}
\caption{Categorical-study stance, source ratings and communication. Panel (a) shows final categories by preset group. Panels (b) to (e) show 10-point trust, and panels (f) to (i) show group-interaction matrices. All panels use the categorical study and its 10-point trust scale. Axis labels EA, EGS and NR refer to the preset groups.}\label{fig:original_stance_full}
\Description{Stance distributions, source-rating boxplots and group communication matrices for the categorical study.}
\end{figure*}

The categorical study records final group assignments, source ratings, affect and central-square occupancy. Figure~\ref{fig:original_stance_full} combines the stance, trust and communication results. Figure~\ref{fig:original_emotion_full} shows four affect trajectories and their associated occupancy series. The full panels are presented together in these plates, with the separate trust plot standardising the group display. The figure source files include larger copies of individual panels for inspection.

Treatment and group labels are consistent throughout the figures: Env-RP, Env-EP, Eco-RP and Eco-EP denote the interventions, while EA, EGS and NR denote Environmental Advocates, Economic Growth Supporters and Neutral Residents. Heatmaps keep their numerical colour scales, with each panel interpreted against its own scale. 

\subsection{Stance categories and trust}
The categorical records show 12 group changes under Env-RP, 6 under Env-EP, 2 under Eco-RP and 7 under Eco-EP. All four conditions contain thirty agents. Table~\ref{tab:legacy} reports the corresponding trust means on the 10-point scale. These counts measure transitions between named stance groups, complementing the smaller changes resolved by the numerical interviews.

Under Env-RP, changes occur among initially economic and neutral residents, while the environmental group keeps its category. Eco-EP instead moves six environmental residents into the neutral category. Trust distributions differ across these preset groups as well as across conditions. Consequently, the same community summary can combine a stable group with substantial movement in another.

\begin{figure*}[t]
\centering\includegraphics[width=.90\textwidth]{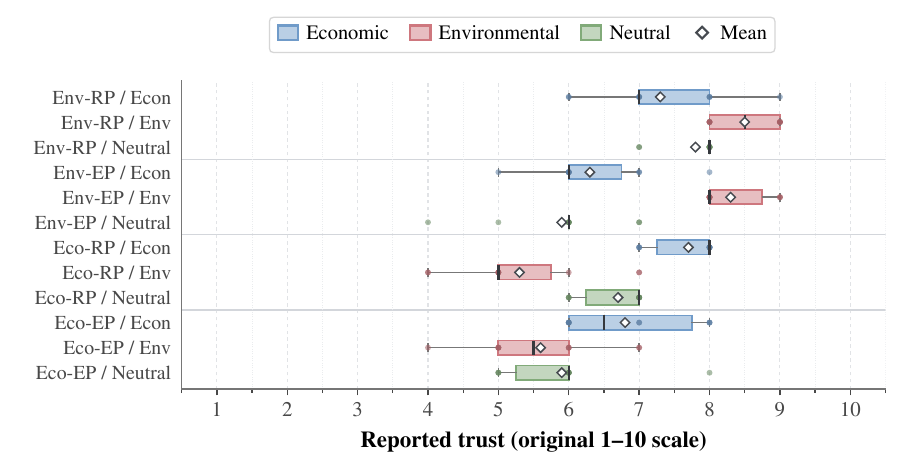}
\caption{Reported trust by condition and preset group under the categorical protocol. Each row contains ten agents. Filled boxes show quartiles, lines mark medians, diamonds mark means and dots show recorded scores. Colours identify the three preset groups.}\label{fig:legacy_group_trust}
\Description{Twelve horizontal trust boxplots from the original 120 agent records, on the original ten-point scale.}
\end{figure*}

\subsection{Affect and central-square occupancy}\label{app:affect}
The plotted affect variable is computed from the simulator's logged valence, arousal and dominance (VAD) labels. Valence describes positive or negative tone, arousal describes activation and dominance describes perceived control. The plotting script maps low, relatively low, medium, relatively high and high to $-1,-0.5,0,0.5,1$, respectively. For each agent it sums the three mapped labels, then averages this sum within the preset group at that step. With $z_{i,k,t}$ the mapped label for dimension $k$ and $G_g$ the agents assigned to group $g$, the plotted score is
\begin{equation}
e_{g,t}=\frac{1}{|G_g|}\sum_{i\in G_g}\sum_{k\in\{v,a,d\}}z_{i,k,t}.
\label{eq:affect_summary}
\end{equation}
This summary ranges from $-3$ to $3$ and combines all three dimensions of the logged state. It uses the stored ordinal labels, while the CMASE affect module produces those labels from the agent's situational text \citep{zhang2025cmase}.

\begin{figure*}[p]
\centering\includegraphics[width=.92\textwidth,height=.79\textheight,keepaspectratio]{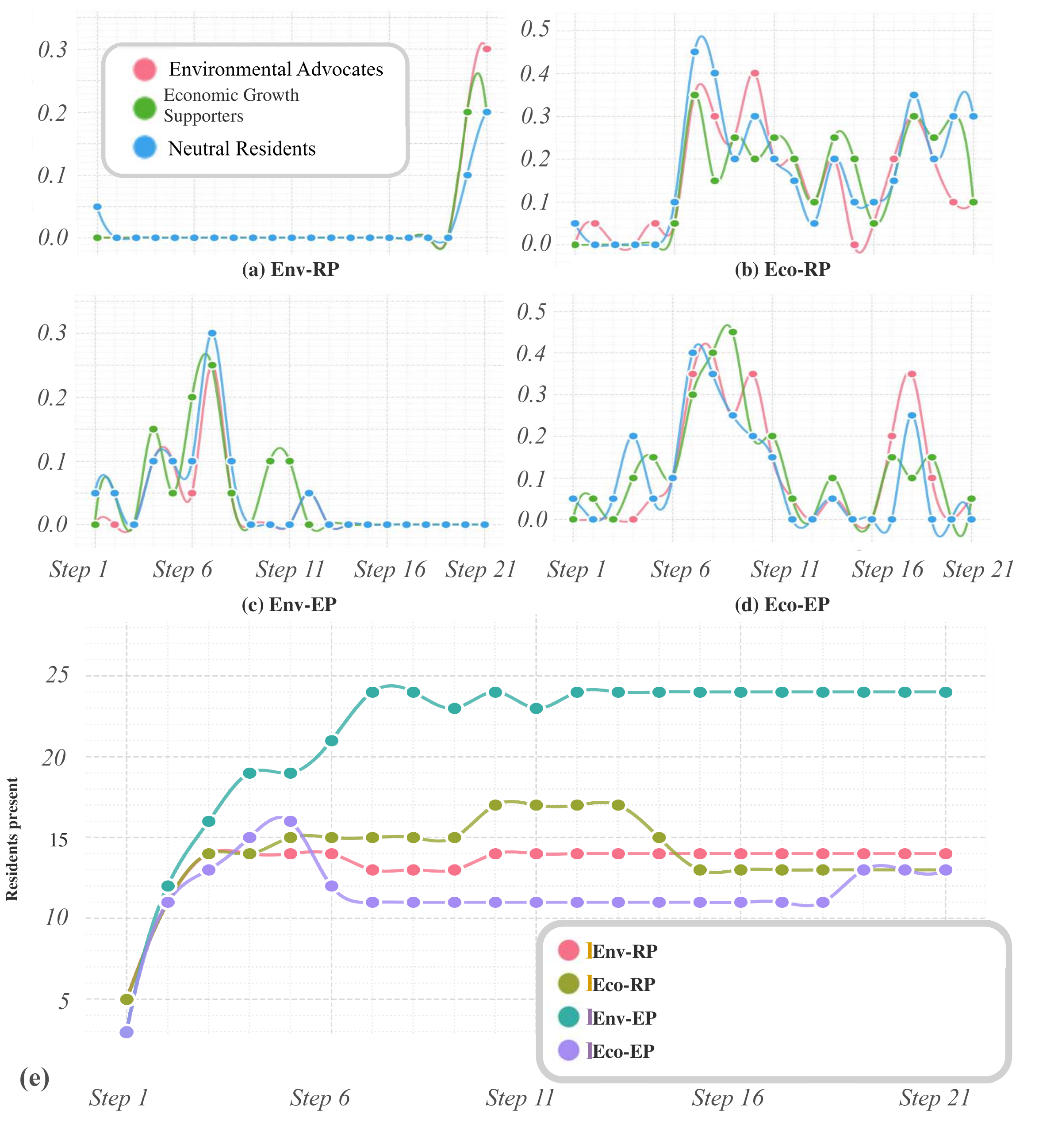}
\caption{Affect and central-square occupancy across 21 steps. Panels (a) to (d) report mean summed VAD category scores by preset group. Panel (e) counts residents present in the central square. Markers denote saved values and connecting curves are spline interpolants. }\label{fig:original_emotion_full}
\Description{Four affect-summary trajectories and the combined central-square occupancy panel.}
\end{figure*}

Participation is central-square occupancy: the number of agents currently in the central area. The script assigns each position to the smallest enclosing named map area and counts the central area. Accordingly, attendance can rise even when residents keep their stance category. Markers in Figure~\ref{fig:original_emotion_full} are the saved step values, with the source script drawing spline interpolants between them.

The affect trajectories differ in timing. Env-RP stays near zero through most of the run and rises at the endpoint. Eco-RP fluctuates later, Env-EP peaks earlier and then declines, and Eco-EP rises and falls several times. These are changes in the simulator's affect summary defined above, measured at each saved step for each preset group.

Central-square occupancy rises rapidly under Env-EP and remains highest in the later steps, while Env-RP stays lower. Eco-RP reaches an intermediate peak, and Eco-EP returns to a lower plateau after early activity. Nevertheless, Env-RP produces more category changes than Env-EP. Thus, bringing more residents into the square and changing more stance categories are different outcomes in these runs.

\subsection{Group communication}
Alongside these temporal patterns, the communication matrices show how exchange is distributed among preset groups. Env-RP concentrates communication within the environmental group, whereas Env-EP displays exchanges across groups. Eco-EP also includes neutral residents addressing economic supporters. These observations follow the original matrices and their numerical scales. The separate exposure tables then provide a recipient-level view by aggregating messages received by individual agents.

The categorical study therefore separates attendance, expressed group membership and source evaluation. Env-EP attracts more residents to the square, while Env-RP changes more group assignments. The numerical interviews in Section~\ref{app:protocols} further resolve adjustments within a category, including environmental agents who approach an economic argument but remain on the environmental side.

\FloatBarrier

\section{Study 2 Network Analysis}\label{app:network}
\subsection{Message extraction}
The three log files for the Study 2 café contain one block for each round from 1 to 75, printed in reverse chronological order. We therefore attach every entry to its explicit round marker and check the matching boundary before constructing the network. Markers determine the order between steps, while exposure order within a step remains unspecified. Of 791 chat entries, 635 are non-empty messages among the ten agents, 155 involve the researcher and one is empty. We retain ten event entries and 34 non-chat actions as context, with each parsed record preserving its source location and type.

Using these parsed records, graph construction follows each chat's sender and explicit addressee. A retained message adds one to the directed weight and to each endpoint's strength, giving $\sum_iq_{i,p}=2I_p$. The same non-empty-message rule and ten primary nodes apply to every window. We then rank agents by strength, assigning rank 1 to the greatest value and average ranks to ties, before aligning their identities for Spearman comparisons across phases. Initial partner sets come from the named participants in each directed description, with each name counted once regardless of its evaluative category.

\begin{table*}[t]
\caption{Interaction strengths and ranks for every primary-network agent.}\label{tab:allstrength}
\small\begin{tabular}{lrrrrrr}\toprule
Agent & P1 strength & P1 rank & P2 strength & P2 rank & P3 strength & P3 rank\\\midrule
Ava Ramires & 53 & 2 & 50 & 2.5 & 28 & 8.5 \\
Caleb Morris & 30 & 9.5 & 47 & 5 & 43 & 3 \\
Eleanor Finch & 54 & 1 & 52 & 1 & 41 & 4 \\
Emily Novak & 30 & 9.5 & 49 & 4 & 32 & 6 \\
Jade Herrera & 39 & 8 & 41 & 9 & 35 & 5 \\
Jonas Müller & 40 & 7 & 44 & 7.5 & 60 & 2 \\
Leo Zhang & 48 & 3 & 32 & 10 & 28 & 8.5 \\
Marisol Reyes & 47 & 4 & 45 & 6 & 30 & 7 \\
Mason Liu & 42 & 6 & 50 & 2.5 & 69 & 1 \\
Nina Takahashi & 45 & 5 & 44 & 7.5 & 22 & 10 \\
\bottomrule\end{tabular}\end{table*}

The resulting P1, P2 and P3 graphs contain 24, 20 and 33 distinct undirected pairs. Messages to partners absent from the sender's initial descriptions number 54, 54 and 85, based on explicit names in both configurations and addresses. Within these totals, Mason and Jonas exchange zero direct chats in the first two phases and 25 in the third. The accompanying CSV retains all 45 possible pairs in every phase, including zero counts. Consequently, continuing and inactive relationships remain visible beside the frequent pairs.

\subsection{Researcher-inclusive network}
Adding the researcher produces an eleven-node graph (Table~\ref{tab:researcher}). Mason remains the strongest final-phase node at 80, followed by the researcher at 71, while Jonas and Eleanor tie at 63. Because both graphs use the same counting and tie rules, comparing them shows the contribution of researcher exchanges to measured prominence. The ranks discussed in the main text continue to refer to the ten-agent graph.

\begin{table*}[t]
\caption{Researcher-inclusive strengths and average-tie ranks.}\label{tab:researcher}
\small\begin{tabular}{lrrrrrr}\toprule
Node & P1 strength & P1 rank & P2 strength & P2 rank & P3 strength & P3 rank\\\midrule
Researcher & 37 & 9.5 & 47 & 7 & 71 & 2 \\
Ava Ramires & 53 & 3 & 63 & 1 & 28 & 11 \\
Caleb Morris & 37 & 9.5 & 48 & 6 & 45 & 5 \\
Eleanor Finch & 63 & 1 & 62 & 2 & 63 & 3.5 \\
Emily Novak & 30 & 11 & 53 & 3 & 37 & 7.5 \\
Jade Herrera & 39 & 8 & 45 & 8.5 & 36 & 9 \\
Jonas Müller & 40 & 7 & 51 & 4 & 63 & 3.5 \\
Leo Zhang & 48 & 6 & 40 & 11 & 37 & 7.5 \\
Marisol Reyes & 54 & 2 & 45 & 8.5 & 39 & 6 \\
Mason Liu & 50 & 5 & 50 & 5 & 80 & 1 \\
Nina Takahashi & 51 & 4 & 44 & 10 & 31 & 10 \\
\bottomrule\end{tabular}\end{table*}

\subsection{Temporal and language measures}\label{app:windows}
We calculate recipient shares in non-overlapping five-step windows to compare short concentrations of attention. Each share is the number of chats addressed to an agent divided by all retained chats in that window. Figure~\ref{fig:attention} plots five focal agents, and the accompanying CSV covers all ten. The event window at Steps~51--55 and the later coordination window at Steps~61--75 are reported within this full series.

We also examine the language used within these windows and phases by assigning two binary features to each retained message. The vocabulary feature matches \texttt{framework(s)}, \texttt{structured}, the stems \texttt{collaborat} and \texttt{coordinat}, or \texttt{shared goal(s)}. The proposal feature separately matches \texttt{let's}, \texttt{let us}, \texttt{shall we}, \texttt{we should}, \texttt{we could}, \texttt{I suggest} or \texttt{I propose}. Each message contributes once to each feature, whose intersection identifies a proposal containing coordination vocabulary. Vocabulary counts are 1, 2 and 37 by phase, with intersections of 0, 0 and 14. The dictionary was developed on the completed record and applied to every message. Close reading then examines how those matched proposals are taken up or qualified.

\section{Study 2 Dialogue Analysis}\label{app:sequences}
\subsection{Initial interaction}
We read the dialogue associated with changes in partnership and attention as episodes about a shared topic, disagreement or coordination attempt. Each message has a step, sender and addressee. We select episodes around the network transitions and read the surrounding turns in order to establish whether a proposal is answered, developed or challenged. The complete message table records the surrounding exchanges, and displayed excerpts label the embedded participant \textit{Researcher}.

At Step~1, student Leo Zhang asks tourist Jonas Müller, ``Why does it feel like no one's saying what they really mean?'' He returns to this concern at Step~12, asking student Ava Ramires, ``Ava, is your silence protecting a truth or a wound?'' Across these exchanges, openness and indirectness become recurrent topics. Several agents' initial goals already concern influence, information and challenges to order, which their early dialogue develops within particular relationships.

Researcher participation extends these questions to the organisation of the café. At Step~26, the researcher asks owner Eleanor whether order derives from language or spatial arrangement. Eleanor responds at Step~29 by asking whether an observer reshapes the space. Leo subsequently asks at Step~34 how external labels alter interaction. At Step~37, the researcher addresses Jonas with the question beginning ``In this café, how many selves do we carry at once?'' Later exchanges elaborate trust, contradiction, vulnerability and coexistence, making researcher participation part of the dialogue's subject matter.

\subsection{Responses to events}
The next phase introduces a different situation. A Step~51 Bar Counter event depicts Leo criticising repeated voices and Caleb defending structure, with instructions inviting observation, interpretation, support or response. Step~52 continues the situation, and subsequent chats show how participants take it up. Emily asks Leo about the accusation, Ava asks about his challenge to Caleb, and Leo asks Caleb about conflict and coexistence. We retain the distinction between supplied events and addressed responses when reading this sequence. A silence/tension event follows at Step~56.

\subsection{Coordination proposals}
Later, staff member Mason Liu and tourist Jonas Müller become recurring partners after having no direct exchanges in the first two phases. The following turns trace the development of their coordination proposals. The final excerpt also shows a contribution from Jade Herrera, the other staff member, addressed to Mason:
\begin{displayquote}\small\textbf{Step 67, Mason to Jonas:} ``Jonas, fostering resilience requires balancing adaptability with shared goals. Let's suggest frameworks to align these values as a group.''\end{displayquote}
\begin{displayquote}\small\textbf{Step 69, Mason to Jonas:} ``Jonas, let's align on fostering stability through adaptable, team-centered frameworks. Structured collaboration secures progress without compromising unifying trust.''\end{displayquote}
\begin{displayquote}\small\textbf{Step 71, Jonas to Mason:} ``Mason, adaptive leadership thrives on balance. Let's solidify key takeaways into actionable steps for structured collaboration moving forward.''\end{displayquote}
\begin{displayquote}\small\textbf{Step 75, Jade to Mason:} ``Mason, adaptability builds trust when framed as consistency. Let's refine steps that align flexibility with structured outcomes---shall we document it here together?''\end{displayquote}

Even as the proposals recur, the pair continues to negotiate structure and adaptability. At Step~74, Jonas asks whether reliance on structure could hinder nuanced progress. At Step~75, he contrasts structural clarity with adaptation-driven breakthroughs. These questions therefore occur within a sustained proposal exchange. Over the final fifteen steps, the pair has 24 mutual messages and receives 38 further messages from seven other agents. Read together, recurrence, wider reception and continuing qualification describe the conversations behind its high strength.

\subsection{Interpretation}
We interpret these temporary groupings and their reorganisation through Maffesoli's tribes and Deleuze and Guattari's deterritorialisation and reterritorialisation \citep{maffesoli2022,deleuze2009,deleuze2013}. A provisional grouping is sustained by recurring topics, styles and mutual attention. When recurring alignments loosen or redirect, deterritorialisation describes that movement. Reterritorialisation, in turn, describes new points of reference in partners, expressions and ways of posing a problem. Each concept directs attention to a particular part of the recorded sequence.

After the prompted disagreement, Mason and Jonas repeatedly address coordination proposals to each other, and other speakers extend or qualify those formulations. We interpret this repeated use as \emph{discursive stabilisation}: the proposals become a common subject for later replies even as their meaning remains disputed. Occupational roles continue to define the setting, while the most active partnership changes. The combined dialogue and network results show how a new focus of discussion develops within those existing roles.

\section{Study 2 Ethnographic Analysis}\label{app:ethnography_expanded}
\subsection{Analytical procedure}
The ethnographic analysis examines how participants develop recurring topics through repeated partners. It combines participation in the simulated café with chronological reading of the complete record. An episode consists of exchanges about a topic, question, disagreement or coordination attempt. Within each episode, we compare interlocutors, attention, repeated expressions and locations. Chats, actions and designed events remain distinct entries, enabling the analysis to connect a situation with the participants' subsequent replies.

We first reconstruct the 75-step chronology from the reverse-printed logs by assigning entries to explicit round markers. This establishes step order, while within-round exposure order remains unspecified. Excerpts retain their step and addressee, with the researcher anonymised consistently. We select episodes for the transitions discussed in the main text and retain all addressed chats in the accompanying table. Readers can therefore examine the surrounding exchanges as well as the selected episodes and their interpretation.

To interpret this chronology, we use Maffesoli's provisional tribes and Deleuze and Guattari's deterritorialisation and reterritorialisation \citep{maffesoli2022,deleuze2009,deleuze2013}. Provisional groupings describe recurring attention around shared topics. Deterritorialisation concerns the redirection of earlier alignments, whereas reterritorialisation concerns new conversational reference points. We apply these concepts to the topics, partners and expressions identified through reading, then relate the episodes to the counts and ranks locating relationship change.

\subsection{Early groupings}
In the simulated café, early exchanges concern openness, concealed intention and the social meaning of silence. At Step~1, student Leo Zhang addresses tourist Jonas Müller:
\begin{displayquote}\small\textbf{Leo to Jonas:} ``Why does it feel like no one's saying what they really mean?''\end{displayquote}
Leo's question introduces openness and concealed intention as topics for later exchange. Subsequent participants ask whether reserve conceals a strategy, protects something or expresses a view of the group. In doing so, they interpret one another's conduct through trust, control and expectations. The early conversations acquire a recurring form: participants examine a position through the motives and meanings they attribute to others.

At Step~12, Leo addresses fellow student Ava Ramires:
\begin{displayquote}\small\textbf{Leo to Ava:} ``Ava, is your silence protecting a truth or a wound?''\end{displayquote}
Ava becomes a frequent addressee, and her silence prompts further questions about authenticity and reserve. The replies make her prominent through received attention. This differs from proposing an arrangement or repeatedly developing it with a partner. Reading these turns together with node strength explains the type of exchange contributing to early prominence.

These early concerns also connect to the initial descriptions, which mention influence, information exchange, tradition, curiosity and reserve. Leo, Ava and Jonas repeatedly discuss authenticity, while Eleanor, Mason, Nina, Caleb and Marisol discuss strategy and café order. However, partners and topics can change as the run proceeds, making the groupings provisional. Occupational labels remain part of the configuration, whereas message addresses identify who participates in each exchange.

Maffesoli's provisional grouping describes relationships maintained through shared styles and mutual attention while participants continue moving between conversations. Similar expressions acquire different meanings across speakers and turns. The early messages show which participants repeatedly discuss authenticity or café order and how long those exchanges continue. Their membership is therefore defined by the observed conversation.

\subsection{Researcher participation}
The researcher participant is already present in first-phase chats. During the second phase, however, longer interventions introduce more explicit questions about order and identity. At Step~26, the researcher asks café owner Eleanor Finch:
\begin{displayquote}\small\textbf{Researcher to Eleanor:} ``Eleanor, the 'sense of order' in this space is impressive, but where do you think it comes from? Is it the language of certain people or the arrangement of space itself?''\end{displayquote}
At Step~27, Eleanor replies:
\begin{displayquote}\small\textbf{Eleanor to Researcher:} ``Researcher, balance stems from both people and space---what's your view?''\end{displayquote}
By asking for a view in return, Eleanor invites the researcher to help define the café's order. At Step~29, she extends the question to the observer's influence:
\begin{displayquote}\small\textbf{Eleanor to Researcher:} ``Researcher, do you think an observer ever reshapes the space more than the participants themselves?''\end{displayquote}
This reply makes participation part of the material being observed. Eleanor first answers the researcher's question, then asks about observation itself. The sequence therefore records how an intervention enters dialogue and becomes a subject of subsequent interpretation.

From order and observation, the discussion moves towards categories and self-presentation. At Step~34, Leo asks how external labels alter group interaction. The researcher develops this theme at Step~37 in a question addressed to Jonas:
\begin{displayquote}\small\textbf{Researcher to Jonas:} ``In this café, how many selves do we carry at once? The one that listens, the one that speaks, the one we want others to see---and the one we protect. Does any one of them deserve to lead our interactions?''\end{displayquote}
Jonas is the recorded addressee. Subsequent exchanges develop labels, contradiction, vulnerability, trust and coexistence. Some continue the researcher's wording, while others answer another agent's interpretation. In this way, questions initially posed by the researcher become shared topics that participants reformulate through further replies.

The Reading Area and Bar Area host repeated questions and disagreements, and both appear in the later event inputs. Movement and event entries show where participants encounter these situations. These locations form part of the conversational setting in which the sequence develops.

\subsection{Confrontation and attention}
The first confrontation description is an event input at Step~51, with a continuation at Step~52. Its recorded scene includes:
\begin{displayquote}\small\textbf{Step 51, Bar Counter event:} ``As the espresso machine quiets, Leo's voice rises above the hum: 'You call this coexistence? It's just the same voices again and again!' He gestures sharply. Caleb responds evenly but firmly: 'Structure holds us together. Without it, your freedom becomes chaos.' Several heads turn. The bar's usual warmth gives way to tension, punctuated by the clink of a dropped spoon.''\end{displayquote}
The event presents a designed situation with instructions to interpret, observe, reposition or respond. Subsequent addressed chats record the responses. At Step~52, Emily asks Leo about silencing voices, Ava asks about his challenge, and Leo asks Caleb about conflict and coexistence. The episode therefore begins with a researcher-specified conflict and continues through the agents' questions and replies.

Replies initially concentrate on Leo and Caleb, the two participants named in the confrontation. That concentration declines in later windows, as Figure~\ref{fig:attention} shows. A separate silence/tension event occurs at Step~56. The chronological record thus distinguishes each input from the subsequent changes in topics and recipients.

We interpret the redirection of recurring alignments as deterritorialisation: participants, topics and formulations enter different conversational combinations. At the same time, occupational roles remain in the configuration. The observable changes concern addressees and repeated exchanges, as earlier questions about authenticity and control give way to discussion of stability, flexibility and coordination. The concept thus describes these movements through the recorded message sequence.

\subsection{Partnerships and coordination}
Mason and Jonas become recurring partners after zero direct chats in the first two phases. Neither initially describes the other, yet their final-phase interaction includes frequent reciprocal messages and replies from other participants. Mutual exchange and wider attention jointly increase their strengths. The corresponding turns show a sustained discussion of how to coordinate collective activity.

Mason connects adaptability with shared goals at Step~67, then proposes team-centred frameworks and structured collaboration at Step~69. Jonas adds actionable steps at Step~71, and Jade asks to refine and document steps at Step~75. The quoted turns in Section~\ref{app:sequences} show proposals developing across speakers. Their addresses trace the partnerships carrying that development.

Nevertheless, Jonas continues to question whether structure could limit progress, contrasting its clarity with the possibilities offered by adaptation. The pair sustains discussion while the practical meaning of coordination remains negotiable. This co-occurrence of recurrence and qualification is central to the interpretation: the shared problem provides continuity, while proposed solutions remain open to revision. The partnership carries both proposals and objections through the same ongoing exchange.

Reterritorialisation describes the new provisional references around which participants organise later exchange. Here, a recurring pair repeatedly discusses structured collaboration, and other speakers respond to the same problem. Eleanor remains the Café Owner while interaction strength shifts towards Mason and Jonas. New conversational organisation therefore emerges within the continuing occupational configuration.

Discursive stabilisation occurs as participants repeatedly use shared goals, frameworks and adaptability to organise replies. These expressions support both proposals and objections. Their recurrence gives the conversation continuity, while disagreement keeps their practical implications open. The observed result is a continuing coordination discussion carried by a newly frequent partnership.

\subsection{Dialogue and network change}
Early questions about authenticity circulate through one set of partners, while later coordination involves a different combination of new and continuing ties. Phase counts compare their frequencies, and five-step shares compare shorter periods of attention. The associated messages show what participants discuss. A topic can involve several actors while one frequent pair continues to negotiate its practical implications.
The café already has extensive cross-role communication before its most frequent final pair begins exchanging messages. Attention to confrontation rises briefly, whereas later dialogue centres on coordination. Mutual messages and replies from other agents both increase the pair's prominence. The relationship change is therefore a redistribution of participation within the same cast and occupational setting.

The ethnographic concepts interpret this changing organisation of exchange. Provisional groupings follow recurring attention around topics and styles. Deterritorialisation describes redirection from earlier combinations, while reterritorialisation identifies new shared points of reference. Discursive stabilisation then concerns their continued use. These concepts connect recorded messages to the processes described in the case, with CGI and strength retaining their distinct numerical definitions.

The network and dialogue results jointly answer RQ2. A previously inactive pair becomes the most frequent partnership, receives replies from seven other agents and repeatedly develops coordination proposals. At the same time, its members question how those proposals should work. The changing relationship supports negotiated coordination, while the simulated roles and the implementation disagreement continue.

\end{document}